%% file: example_paper.tex
\documentclass{article}

\usepackage{microtype}
\usepackage{graphicx}
\usepackage{subfigure}
\usepackage{booktabs} 

\usepackage{hyperref}

\usepackage[accepted]{icml2024}

\usepackage{amsmath}
\usepackage{amssymb}
\usepackage{mathtools}
\usepackage{amsthm}

\usepackage[capitalize,noabbrev]{cleveref}

\theoremstyle{plain}

\theoremstyle{definition}

\theoremstyle{remark}

\usepackage[textsize=tiny]{todonotes}

\newcommand{\dprobe}[0]{$\mathcal{D}_{\text{probe}}$}

\icmltitlerunning{Linear Explanations for Individual Neurons}

\begin{document}

\twocolumn[
\icmltitle{Linear Explanations for Individual Neurons}


\icmlsetsymbol{equal}{*}

\begin{icmlauthorlist}
\icmlauthor{Tuomas Oikarinen}{cse}
\icmlauthor{Tsui-Wei Weng}{hdsi}
\end{icmlauthorlist}

\icmlaffiliation{cse}{CSE, UC San Diego, CA, USA}
\icmlaffiliation{hdsi}{HDSI, UC San Diego, CA, USA}

\icmlcorrespondingauthor{Tuomas Oikarinen}{toikarinen@ucsd.edu}
\icmlcorrespondingauthor{Tsui-Wei Weng}{lweng@ucsd.edu}

\icmlkeywords{Machine Learning, ICML}

\vskip 0.3in
]

\printAffiliationsAndNotice{}  

\begin{abstract}
In recent years many methods have been developed to understand the internal workings of neural networks, often by describing the function of individual neurons in the model. However, these methods typically only focus on explaining the very highest activations of a neuron. In this paper we show this is not sufficient, and that the highest activation range is only responsible for a very small percentage of the neuron's causal effect. In addition, inputs causing lower activations are often very different and can't be reliably predicted by only looking at high activations. We propose that neurons should instead be understood as a linear combination of concepts, and develop an efficient method for producing these linear explanations. In addition, we show how to automatically evaluate description quality using \textit{simulation}, i.e. predicting neuron activations on unseen inputs in vision setting.
\end{abstract}

\input{1_intro}

\input{2_activation_importance}

\input{3_method}

\input{4_simulation}
\input{5_result}

\input{6_related_work}

\input{7_conclusion}

\clearpage

\section*{Acknowlegements}
This work is supported in part by National Science Foundation (NSF) awards CNS-1730158, ACI-1540112, ACI-1541349, OAC-1826967, OAC-2112167, CNS-2100237, CNS-2120019, the University of California Office of the President, and the University of California San Diego's California Institute for Telecommunications and Information Technology/Qualcomm Institute. Thanks to CENIC for the 100Gbps networks. T. Oikarinen and T.-W. Weng are supported by National Science Foundation under Grant No. 2107189 and 2313105. T.-W. Weng also thanks the Hellman Fellowship for providing research support.

\section*{Impact statement}
Our paper proposes improved methods to understand neural networks, and as such we expect it's impact on society to be mostly positive. While our method is only a small part of this, we believe better understanding of networks reduces the chances of harm caused by deploying unsafe or unreliable models in important settings. One potential downside of explanations is that they risks giving the illusion of understanding, without actually being faithful to the model. We hope to have reduced the chances of this with significant focus on better evaluations of neuron explanations.

\clearpage
\newpage
\bibliography{example_paper}
\bibliographystyle{icml2024}


\input{A_Appendix}
\input{B_Appendix_Ablations}
\input{C_Appendix_more_results}


\end{document}

%% file: 1_intro.tex
\section{Introduction}
\label{sec:intro}

Current machine learning models are extremely capable at many tasks, yet they are notoriously hard to understand, and most often seen as black boxes. Recently, the field of \textit{mechanistic interpretability} \cite{olah2020zoom, elhage2021mathematical} has emerged to address this issue by providing mechanistic understanding of the inner workings of neural networks, with the ultimate goal of reverse engineering the algorithms that neural networks use to solve problems. 

Individual neurons (or channels of a CNN) are the most simple unit of a neural network, and understanding them is a fundamental building block of mechanistic interpretability. Many methods have been developed to understand individual neurons, based on both manual inspection \cite{erhan2009visualizing, zhou2014object, olah2020zoom} and automated description \cite{netdissect2017, hernandez2022natural, oikarinen2023clip}. However, most of these methods only focus on understanding and explaining the very highest activations of a neuron. In Section \ref{sec:act_importance} we show that this is not enough, and a neuron's impact on network outputs is distributed quite evenly across all inputs. Therefore, to fully understand a neuron, we need to understand everything it is doing, not just its highest activations.

In Section \ref{sec:method} we propose a solution to these issues, which we call \textbf{Linear Explanations (LE)}. In our method, each neuron is described as linear combination of concepts, such as "$w_1 \times$ dog + $w_2 \times$ pet". This explanation allows us to directly simulate neuron activations as $s(x_i) = w_1 \mathbb{P}(\text{dog}|x_i) + w_2 \mathbb{P}(\text{pet}|x_i)$, where $\mathbb{P}(\text{concept}|x_i$) can be estimated by either a human or a model. Our automated method allows for learning accurate linear explanations efficiently, as demonstrated in 
Section \ref{sec:result}, where we evaluate and compare our method and existing automated explanation methods. 
We believe linear explanations are a natural way to represent neurons, and they can elegantly model the scalar nature of a neuron's activation. In addition, linear explanations can be found efficiently and do not binarize neuron activations (which loses information), unlike alternative methods to capture complex neuron behavior such as Compositional Explanations \cite{mu2021compositional}.

In Section \ref{sec:simulation} we propose a new, more rigorous way to automatically evaluate the quality of neuron descriptions via \textit{simulation}, i.e. predicting the neuron's activation on a new input given the explanation. The simulation evaluation was recently proposed for explaining neurons of large language models by \cite{bills2023language}, and has since been a popular evaluation method in language settings \cite{cunningham2023sparse, bricken2023monosemanticity}. We are the first to present an effective and natural way to run simulation on vision models by utilizing SigLIP \cite{zhai2023sigmoid} models as the simulator. We find that existing neuron explanations~\cite{netdissect2017, hernandez2022natural, oikarinen2023clip} perform poorly under simulation, while our method provides more than $3 \times$ higher ablation scores when simulated. Finally, in Figure \ref{fig:area_chart} we show how our method can uncover multiple roles played by the same neuron that would be missed when only looking at highest activations. Our code and results are available at \href{https://github.com/Trustworthy-ML-Lab/Linear-Explanations}{https://github.com/Trustworthy-ML-Lab/Linear-Explanations}.

%% file: 2_activation_importance.tex
\section{Motivation: How Important are Different Parts of a Neuron's Activation Pattern?}
\label{sec:act_importance}

In this section, we aim to answer the following question:

\textit{Is most of a neuron's impact on the network caused by the very highest activating inputs, or are all inputs important?}

We do this by ablating the individual neurons of a network, i.e. replacing their activation by 0, and measuring the change in network outputs. We refer to this as the neuron's causal effect. 

\subsection{Definitions}

To describe our results, we first need to define a few metrics:
The idea is to measure the causal effect of these neurons in terms of metrics most relevant to the end use case, which in a classification setting are accuracy and cross-entropy loss. 

Let $f(\cdot)$ be a neural network of interest. $f(x)$ is the network's output on an input $x$, i.e. class probabilities. Let $D = \{(x_i, y_i)\}$ be a dataset with images $x_i$ and corresponding ground-truth class labels $y_i$. We define $I_k(x_i, y_i)$ as the impact of neuron $k$ on the input $x_i$:
\begin{equation}
    I_k(x_i, y_i) :=  \frac{\Delta \text{Acc}_k(x_i, y_i) - \Delta L_k(x_i, y_i)}{2}
\end{equation}
where $\Delta \text{Acc}_k(x_i, y_i)$ is the change in accuracy and $\Delta L_k(x_i, y_i)$ is the change in loss as formally defined below:
\begin{equation}
    \Delta \text{Acc}_k(x_i, y_i) = \frac{[h({f}(x_i)) = y_i] - [h({f}_{\sim k}(x_i)) = y_i]}{\sum_{(x_j, y_j) \in D}[h(f(x_j)) = y_j]}
\end{equation}
\begin{equation}
    \Delta L_k(x_i, y_i) = \frac{L(f(x_i), y_i) - L(f_{\sim k}(x_i), y_i)}{\sum_{(x_j, y_j) \in D} L(f(x_j), y_j)}
\end{equation}
Here $[\cdot]$ is the indicator function, taking value 1 if the expression is True and 0 otherwise, $h(\cdot) \coloneqq \text{argmax}(\cdot)$, $f_{\sim k}(\cdot)$ is the output of the model without neuron $k$, i.e. the neuron's activation is replaced by 0 and $L$ is the loss function, such as cross-entropy loss. We use temperature calibration \cite{guo2017calibration} before calculating the losses.

The impact $I_k(x_i, y_i)$ is then the average of the neuron's effect on accuracy and loss of the model on an input, as a fraction of the model's total loss and accuracy, with signs chosen such that a positive impact means including the neuron $k$ improves the network's predictions.

Next, we measure how much of a neuron's total impact is caused by the inputs that activate it the highest. 
To achieve this, we define \textit{Top Impact}, denoted as $TI_k(\beta)$, with the input argument $\beta \in [0,1]$ representing the fraction of highest activating inputs included:
\begin{equation}
    TI_k(\beta) = \frac{\sum_{i=1}^{\beta|D|} |I_k(x_{(i)}, y_{(i)})|}{\sum_{(x_j, y_j) \in D} |I_k(x_{j}, y_{j})|}
\end{equation}
where $x_{(i)}, y_{(i)}$ are ordered in descending order of neuron $k$'s activation. I.e. $g(A_k(x_{(i)})) \geq g(A_k(x_{(i+1)})) \, \forall i$, where $A_k(x_i)$ is the activation of a neuron or a channel in CNNs on input $x_i$, and $g$ is a summary function such as mean or max that takes the 2D activation map of a CNN channel into a single scalar (or identity for scalar neurons) as defined in~\cite{oikarinen2023clip}. A neuron where all inputs are equally important should have $TI_k(\beta) = \beta$.

\begin{figure*}[t!]
    \centering
    \includegraphics[width=0.8\linewidth]{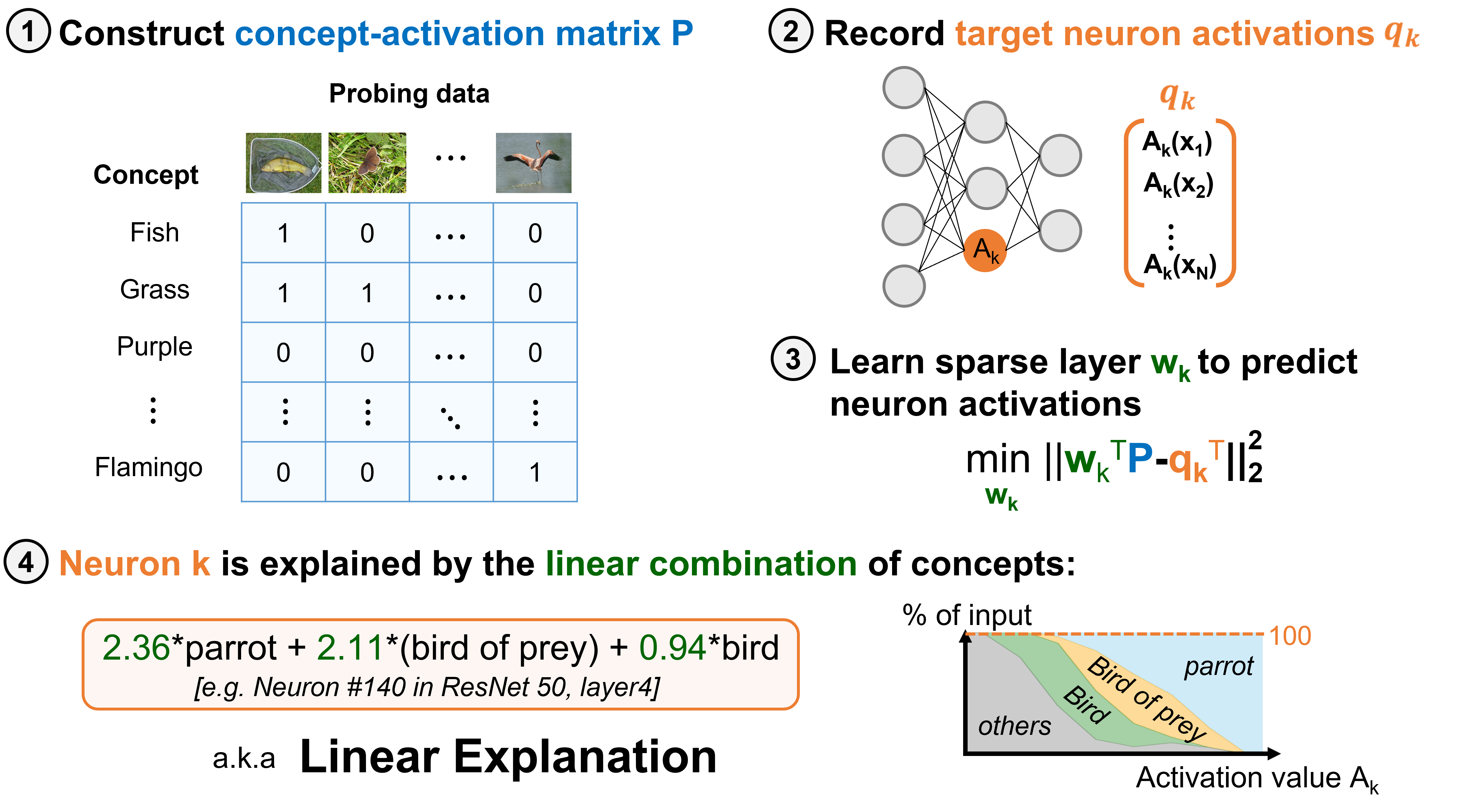}
    \caption{Overview of our proposed method: Linear Explanations.}
    \label{fig:overview}
\end{figure*}

\subsection{Results}

\input{table/top_impact}

To measure how important highly activating inputs are to network predictions in practice, we ablated out all the neurons (channels) of ResNet-50 \cite{he2016deep} one at a time and measured the change in performance on ImageNet validation data. Results in terms of \textit{Top Impact} are shown in Table \ref{tab:top_impact}. We used mean as summary function $g$.
We can see that if we wish to understand most of a neuron's impact (i.e. high $TI$), we need to look at a large fraction of the neurons inputs (high $\beta$), not just the most highly activating inputs. 

This is in contrast to many popular methods of single neuron explanation, which often focus exclusively on the most highly activating inputs, i.e. very small $\beta$. For example MILAN \cite{hernandez2022natural} only looks at 15 most highly activating inputs, which is equivalent to $\beta = 0.0003$. In Table \ref{tab:top_impact}, we can see that these inputs only explain 0.258\% of the neuron's impact on average. Similarly, CLIP-Dissect \cite{oikarinen2023clip} with soft-wpmi activation function only looks at 100 ($\beta = 0.002$) most highly activating inputs, which make up 1.522\% of the neuron's impact. 
While Network Dissection \cite{netdissect2017} looks at all inputs, it only aims to explain top 0.5\% of location specific neuron activations which causes similar issues. Based on the findings in Tab~\ref{tab:top_impact}, we believe most inputs of a neuron are important, and only focusing on highest activating inputs (small $\beta$) is not sufficient to faithfully explain individual neurons, or to evaluate how good such explanations are. To resolve this issue, we propose a new explanation method (Section \ref{sec:method}) and a new evaluation metric (Section \ref{sec:simulation}) that focus on explaining the \textit{entire} range of a neuron's activations.

%% file: table/top_impact.tex
\begin{table}[b!]
\scalebox{0.82}{
\begin{tabular}{@{}l|llllll@{}}
$\beta$ & Conv 1 & Layer 1 & Layer 2 & Layer 3 & Layer 4 & All \\ \midrule
0.0003 & 0.054\% & 0.084\% & 0.110\% & 0.268\% & 0.319\% & 0.258\% \\
0.002 & 0.319\% & 0.454\% & 0.574\% & 1.369\% & 2.007\% & 1.522\% \\
0.02 & 2.743\% & 3.526\% & 4.160\% & 8.654\% & 13.59\% & 10.22\% \\
0.1 & 12.42\% & 14.80\% & 16.53\% & 28.28\% & 39.38\% & 31.42\% \\
0.5 & 55.23\% & 59.49\% & 62.08\% & 76.85\% & 84.55\% & 77.46\% \\ \bottomrule
\end{tabular}
}
\caption{Average \textit{Top Impact} for neurons in different layers of ResNet-50 (ImageNet). We can see that while highly activating inputs (small $\beta$) are more impactful than the average input, especially on later layers, they still explain only a small portion of the neuron's total effect. E.g. $\beta = 0.002$ only accounts for around 1.52\% of Total Impact.}
\label{tab:top_impact}
\end{table}

%% file: 3_method.tex
\section{Method}

\label{sec:method}

In this section we present our method \textbf{Linear Explanations} or \textbf{LE}. Our method consists of two main parts: (i) Constructing a concept activation matrix $P$, and (ii) Learning a linear combination to explain the neuron. An overview of our method is shown in Figure \ref{fig:overview}. 

\subsection{Constructing a concept activation matrix}

An essential part of our explanation is creating the concept activation matrix $P$. Each entry $P_{ij}$ represents how much of concept $c_i$ is present in input $x_j$ in a way that is aligned with human perception. That is $P_{ij} \approx \mathbb{P}(c_i|x_j)$, for each input $x_j$ of the probing dataset $\mathcal{D}_{\text{probe}}$($|\mathcal{D}_{\text{probe}}| = N$) and each concept $c_i$ in the concept set $\mathcal{S}$($|\mathcal{S}|=M$). 

In this paper we experiment with two different methods of determining $P$ described below: \textbf{Label} and \textbf{SigLIP}. 

\textbf{Label: Use labeled data.} 
If we have access to labeled data for all concepts in the concept set for all inputs in the probing dataset, we can directly use these labels as our concept activation matrix $P$. This is the ideal case and works in situations where such data is available, such as the Broden~\cite{netdissect2017} dataset or the class labels on a validation dataset. We denote this approach as \textbf{LE(Label)}.
However, we often do not have access to labels for all concepts and/or images we want to utilize, and collecting additional labeled data can be expensive. This leads us to our second method \textbf{SigLIP}.

\textbf{SigLIP: Pseudo-labels from Multimodal Models.}
In cases where sufficient labeled data is not available, it is often beneficial to create artificial labels using pretrained foundation models such as CLIP \cite{radford2021learning}. In our experiments we used SigLIP \cite{zhai2023sigmoid}(ViT-L-16-384) as our \textit{explainer model}. SigLIP(ViT-L-16-384) is a more recent model similar to CLIP, and was chosen because of its improved performance, as well as use of sigmoid function during pretraining. Training with sigmoid activation is important for creating a reliable concept activation matrix $P$, as it is essentially a multi-label classification task, i.e.  most images contain multiple concepts. Because of this, softmax based models will likely perform poorly on constructing a concept activation matrix $P$. 

To ensure the pretrained SigLIP model is aligned with human concept predictions, we add additional calibration parameters $a$ and $b$ and optimize these using publicly available data. Let $E_I$ be the image encoder and $E_T$ be the text encoder of the SigLIP model. We generate concept activation matrix $P$ as follows: $P_{ij} = \sigma (a \cdot E_T(c_i) E_I(x_j) + a \cdot b)$. Since our goal is to have $P_{ij} \approx \mathbb{P}(c_i|x_j)$, i.e. to be similar to human predictions, we learn $a$ and $b$ by minimizing binary cross-entropy loss $L_{BCE}$ on concepts we have labels for. In particular we use ImageNet validation data, with added superclass labels according to WordNet hierarchy~\cite{miller1995wordnet}, which turns this into a multi-label classification task. 
$a$ and $b$ are then determined as: 
\begin{equation}
    \min_{a, b} \sum_{x_i \in D} \sum_{c_j \in C} L_{BCE}(\sigma (a \cdot E_T(c_j) E_I(x_i) + a \cdot b), y_{i, j})
    \label{eq:a_b_tuning}
\end{equation}
where $C$ is the set of [super]class names, $c_j$ is the name of $j$-th [super]class, $y_{i, j}$ is a binary label indicating whether input $x_i$ belongs to [super]class $c_j$ and $L_{BCE}$ is binary cross-entropy loss. The hyperparameters $a$ and $b$ are explainer model specific, and independent of $\mathcal{D}_{\text{probe}}$ or $\mathcal{S}$.


\subsection{Learning Linear Explanations}

Once we have our concept activation matrix $P \in \mathbb{R}^{M \times N}$, our next task is to learn a sparse linear model that can predict the activation of our target neuron based on the presence of a few concepts in our input. We do this in two steps described below.
Let $q_k \in \mathbb{R}^{N}$ be the activation vector of neuron $k$, defined as 
$q_k = \begin{bmatrix}
    g(A_k(x_1)) \\
    g(A_k(x_2)) \\
    \vdots \\
    g(A_k(x_N))
\end{bmatrix}$, where $A_k(x_i)$ is the activation of neuron $k$ on input $x_i$, and $g(\cdot)$ is a summary function(mean) which is used in case the neuron's activation is not a scalar, for example when neuron $k$ is a channel of a CNN.

Throughout the paper we use a 70-10-20 split to divide our $\mathcal{D}_{\text{probe}}$ into train, validation and test set splits to avoid overfitting our explanations. We denote the train subset of $q_k$ and $P$ as $q_k^{\text{train}}$ and $P^{\text{train}}$ respectively.

\subsubsection{Learn a relatively sparse $w_k$}
First, we use the GLM-Saga package \cite{wong2021leveraging} to learn a sparse linear weight $w_k \in \mathbb{R}^{M}$ to minimize the following objective:
\begin{equation}
\label{eq:L_MSE}
    \mathcal{L}_{\text{MSE}} = ||w_k^\top P^{\text{train}} - (q_k^{\text{train}})^\top||_2^2 + \lambda R_{\eta}(w_k)
\end{equation}
where $R_{\eta}(w_k) = (1-\eta)\frac{1}{2}||w_k||_2^2 + \eta ||w_k||_{1}$ and $\eta$ is a hyperparameter, set to 0.99 in our experiments.  It is worth noting that we optimize the predictions to be accurate on the entire $\mathcal{D}_{\text{probe}}^{\text{train}}$ (i.e. $\beta = 1$), so our explanations describe the entire activation range, not just the most highly activating inputs as discussed in Sec~\ref{sec:act_importance}. Our goal is to learn $w_k$ that can accurately predict neuron activations based on just the concept information embedded in the concept activation matrix $P$, while also being sparse for interpretability. The sparsity constraint is reflected in the term $||w_k||_{1}$ of the regularization $R_{\eta}(w_k)$. This is a surrogate term for the exact sparsity goal of minimizing the $\ell_0$ norm (i.e. $||w_k||_{0}$), as the $\ell_1$ norm is convex and much easier to optimize than the non-convex $\ell_0$ norm. However we found it hard/unstable to find extremely sparse (around 5 concepts per neuron) $w_k$ using only this method. 

In Section \ref{sec:greedy_search} we discuss how to overcome this using greedy search guided by the found values $w_k$. For discussion on results without using greedy search, see Appendix \ref{app:gs_ablation}. 
 


\subsubsection{Greedy search}

\label{sec:greedy_search}

Our basic idea for greedy search is to use the weights $w_k$ found in previous step as a heuristic to find a very sparse explanation for the neuron of interest. This greedy-search process is described in detail in Algorithm \ref{alg:greedy_search} and Appendix \ref{app:greedy_search_alg}.  For each neuron, we test $r=10$ available concepts with the highest weight, and choose the one that can train the best model together with the already selected concepts. We continue this process until we reach the maximum number of concepts $v=10$, or until adding another concept does not improve performance enough. This is modeled by the tolerance parameter $\epsilon$. This lets our method dynamically decide the number of concepts for each neuron, adding more complexity to the explanation only when it is needed. The tolerance parameter can be adjusted to make a tradeoff between simpler and shorter v.s. more complete explanations. This process requires training many models per each neuron, but they are extremely small linear models (10 or less parameters) which can be trained to optimal in a fraction of a second. Each of these models is trained to minimize MSE loss of predicting neuron activations as a linear function of only the selected concepts (with no regularizer), i.e. the first term in Eq~\eqref{eq:L_MSE}. More concretely, given set of concept indices discovered from Greedy search $\mathcal{I} = \{i_1, ..., i_u\}$, $u \leq v$, our final explanation $E$ is then:
\begin{equation}
    E = \{(w^*_{k, 1}, c_{i_1}), \cdots, (w^*_{k, u}, c_{i_u})\}
\end{equation}
where $w^*_k = \text{argmin}_{w \in \mathbb{R}^u} ||w^\top P^{\text{train}}_{\mathcal{I}} - (q_k^{\text{train}})^\top||_2^2$.

%% file: 4_simulation.tex
\section{Improving Evaluation of Explanations via Simulation}

\label{sec:simulation}

\begin{figure*}[t]
    \centering
    \includegraphics[width=0.8\linewidth]{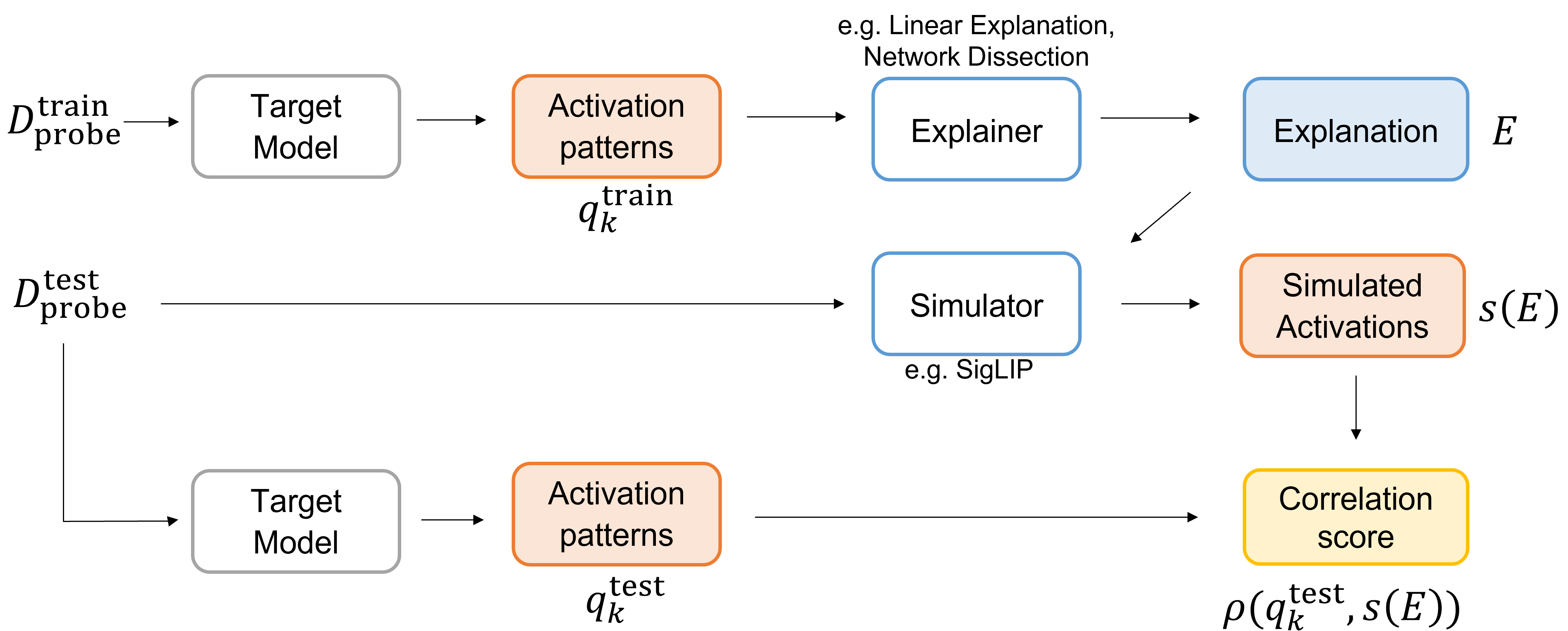}
    \caption{An overview of the simulation pipeline with correlation scoring.}
    \label{fig:simulation_overview}
\end{figure*}

An important part of creating explanations for individual neurons is being able to evaluate how faithfully these descriptions actually correspond to target neuron behavior. This has traditionally been done with methods such as evaluating whether the description matches the most highly activating images \cite{netdissect2017, hernandez2022natural, oikarinen2023clip, kalibhat2023identifying}, but this only evaluates a very small portion of the neuron's activations, which is not sufficient for understanding the neuron as we have shown in Section \ref{sec:act_importance}. In addition, \cite{zimmermann2023scale} showed that while the very highest activations of a neuron may often be understandable, understanding a larger part of the activation range quickly becomes difficult for humans.

Inspired by recent work in language models \cite{bills2023language}, we instead propose to evaluate our explanations using \textit{simulation}. See Figure \ref{fig:simulation_overview} for an overview of our simulation pipeline. The basic idea of \textit{simulation} is as follows: 
\begin{enumerate}
    \item Generate a human understandable explanation $E$ for the neurons using an \textit{Explainer}\footnote{Explainer could be a neural network, a human or an algorithm using human-annotated labels}.
    \item Use a \textit{Simulator} to predict neuron activations $s(x, E)$ on new inputs $x$, based on only the explanation $E$ and the input. 
    \item Score how well the simulated activations fit the actual neuron activations, using either correlation or ablation scoring.
\end{enumerate}
 The intuition behind simulation is that a good explanation should allow the simulator model (or a human) to accurately predict the neuron activations. \cite{bills2023language} used GPT-4 as both the explainer and the simulator, showing the 5 most highly activating text excerpts from a subset of its training data to generate an explanation, and then simulate its activations on 5-10 new excerpts to evaluate how good these explanations are. 

In this section, we propose how to naturally extend the idea of simulation from language to the vision domain, and discuss some conceptual and practical improvements we can do in this modality. We argue a good simulator should have the following properties:
\begin{enumerate}
    \item The simulator needs to be able to take any textual concept, and predict how highly it activates on any given image (or subpart of the image).
    \item The simulator and the explainer should be \textbf{different} models. Using the same model may overestimate the quality of the explanations. For example, the explainer could use an uninterpretable code, or a misunderstood concept to explain a neuron and still receive a high score if the simulator shares this misunderstanding.
    \item Third, a simulator should be \textit{human aligned} i.e. predict activations similar to how a human would.
\end{enumerate}
Following the above, we propose using CLIP \cite{radford2021learning} or similar models to perform this image level simulation. In particular, we choose the SigLIP-SO400M-14-384 \cite{zhai2023sigmoid} as the simulator, because it is the most powerful sigmoid trained model available. Note this is a different SigLIP model than the one we use to generate explanations of \textbf{LE(SigLIP)}. 

To ensure that our simulator is human-aligned, we base our simulation on predicting $\mathbb{P}(c_i|x_j)$, i.e. how likely is the concept represented by $c_i$ is to be present in image $x_j$. This is important as it is aligned with how humans usually think about concepts, instead of directly predicting real values as done by \cite{bills2023language}. We estimate $\mathbb{P}(c_i|x_j)$ similar to how we constructed matrix $P$ in Section \ref{sec:method}, i.e. 
\begin{equation}
    \mathbb{P}_{sim}(c_i|x_j) = \sigma (a_{sim} E_T^{sim}(c_i) E_I^{sim}(x_j) + a_{sim} b_{sim})
\end{equation}
Like before, we optimize the hyperparameters $a_{sim}$ and $b_{sim}$ on ImageNet validation data with superclasses as defined in Eq. \eqref{eq:a_b_tuning}. Different from explainer (Section \ref{sec:method}), the simulator does not use a fixed concept set, but instead evaluates all the concepts present in explanation $E$.


For explainers that produce a single text explanation $c_e$ ~\cite{netdissect2017, hernandez2022natural, oikarinen2023clip}, their explanation can be written as a linear explanation of length 1, i.e. $E = \{(1, c_e)\}$.
Once we have an explanation $E$, the initial simulated activation $s(x_j, E)$ is calculated in the following way: 
\begin{equation}
    s(x_j, E) = \sum_{(w_i, c_i) \in E} w_i \mathbb{P}_{sim}(c_i | x_j)
\end{equation}
With $s(x_j, E)$, we can evaluate the quality of explanation $E$ using two scoring methods:

\begin{figure*}[t]
    \centering
    \includegraphics[width=0.95\linewidth]{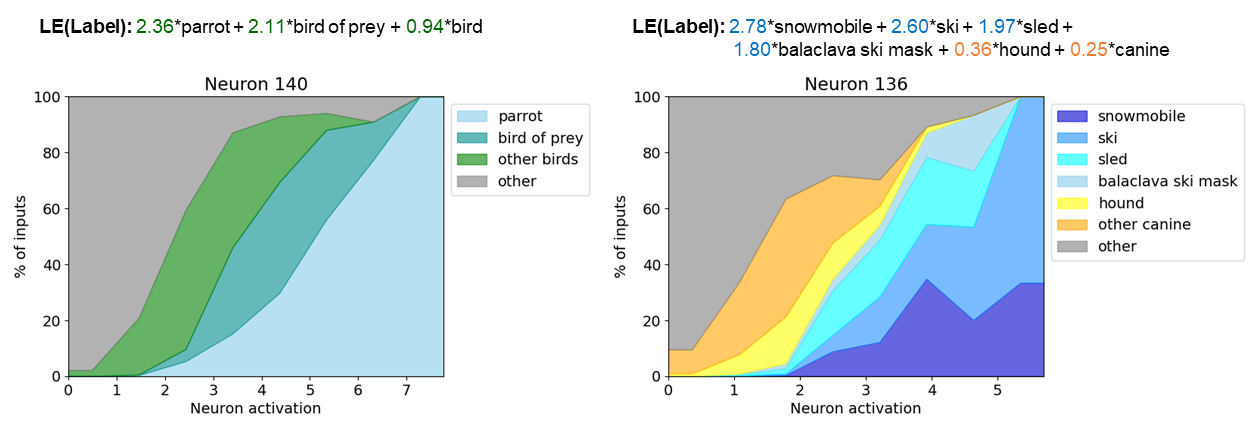}
    \caption{An area chart of the activations of two neurons in layer4 of ResNet-50. We can see Neuron 140 is mostly monosemantic, but represents different types of birds at different activation ranges. In contrast, neuron 136 has two distinct roles, snow and skiing related concepts at high activations and dog-like animals at lower activations.}
    \label{fig:area_chart}
\end{figure*}

\textbf{A. Correlation Scoring $\rho(k, E)$}:  The explanations $E$ are scored based on the correlation coefficient between the simulated activations and the neuron's real activations patterns. 

\begin{equation}
    \rho(k, E) = \sum_{x \in \mathcal{D}_{\text{probe}}^{\text{test}}} \frac{\hat{s}(x, E) \cdot \hat{g}(A_k(x))}{|\mathcal{D}_{\text{probe}}^{\text{test}}|}
\end{equation}
where $\hat{s}$ and $\hat{g}(A_k)$ are $s$ and $g(A_k)$ normalized to have mean 0 and standard deviation of 1 on the test distribution.

\textbf{B. Ablation scoring $\alpha(k, E)$}: In ablation scoring, we directly replace the actual neuron activation with our simulated neuron activations, and measure how much this changes the final model outputs. 
Unlike correlations scoring, for ablation scoring we need to make sure simulated activations have the right scale. This leads us to ablation simulated value $s_{abl}(x, E, c, d)$, where $c$ and $d$ are scaling parameters used to match the magnitude of predicted activations with actual neuron activations:
\begin{equation}
    s_{abl}(x, E, c, d) = c \cdot s(x, E) + d
\end{equation}
We measure ablation performance with an objective adapted for classification setting, defined as $\alpha_{init}(k, \mathcal{D}, E, c, d) =$
\begin{equation}
    1 - \frac{\sum_{(x, y) \in \mathcal{D}} |L(f_{k\leftarrow s_{abl}(x, E, c, d)}(x), y) - L(f(x), y)|}{\sum_{(x, y) \in \mathcal{D}} |L(f_{k\leftarrow \mu}(x), y) - L(f(x), y)|}
\end{equation}

where $f(x)$ is the output of the target model, $L$ is the loss function (e.g. cross-entropy-loss), $f_{k\leftarrow s_{abl}(x, E, c, d)}(x)$ indicates replacing the activations of neuron $k$ with the simulated values $s_{abl}(x, E, c, d)$, while $f_{k\leftarrow \mu}(x)$ means replacing the neuron $k$'s activation with its mean value. This is the same as the ablation objective of \cite{bills2023language}, except we have replaced Jensen-Shannon divergence with cross-entropy loss as it is a more relevant metric in the classification task. Note this formulation requires using $\mathcal{D}$ where we have access to ground truth labels, which is the case when using validation data. 

To get the parameters $c$, $d$ , we optimize with gradient descent on the validation split:
\begin{equation}
    c^*, d^* = \text{arg}\max_{c,d} \alpha_{init}(k, \mathcal{D}_{\text{probe}}^{\text{val}}, E, c, d) 
\end{equation}
Our final simulation score is then evaluated on the test split, and defined as:
\begin{equation}
    \alpha(k, E) = \alpha_{init}(k, D_{\text{probe}}^{\text{test}}, E, c^*, d^*)
\end{equation}
A perfect simulation will reach $\alpha(k, E) = 1$, while random guess should receive a score of 0.

Here we used optimization to find the parameters $c$ and $d$. \cite{bills2023language} instead calculated $c$ and $d$ based on correlation between the predicted and true activation. We evaluate the difference between these choices in Appendix \ref{app:ablation_scaling}, and find our optimization method gives noticeably higher ablation scores, but with a higher computational cost.

Finally we note that our SigLIP based simulation pipeline is much more computationally efficient than the GPT-4 pipeline of \cite{bills2023language} because we do not need to recalculate image embeddings when simulating a new explanation. 
This allows us run simulation on the entire test split of probing data (10,000 images for ImageNet), which is many orders of magnitude larger than the number of samples \cite{bills2023language} used for simulations.

%% file: 5_result.tex
\section{Experiment Results}
\label{sec:result}

\begin{figure*}[]
    \centering
    \includegraphics[width=0.9 \linewidth]{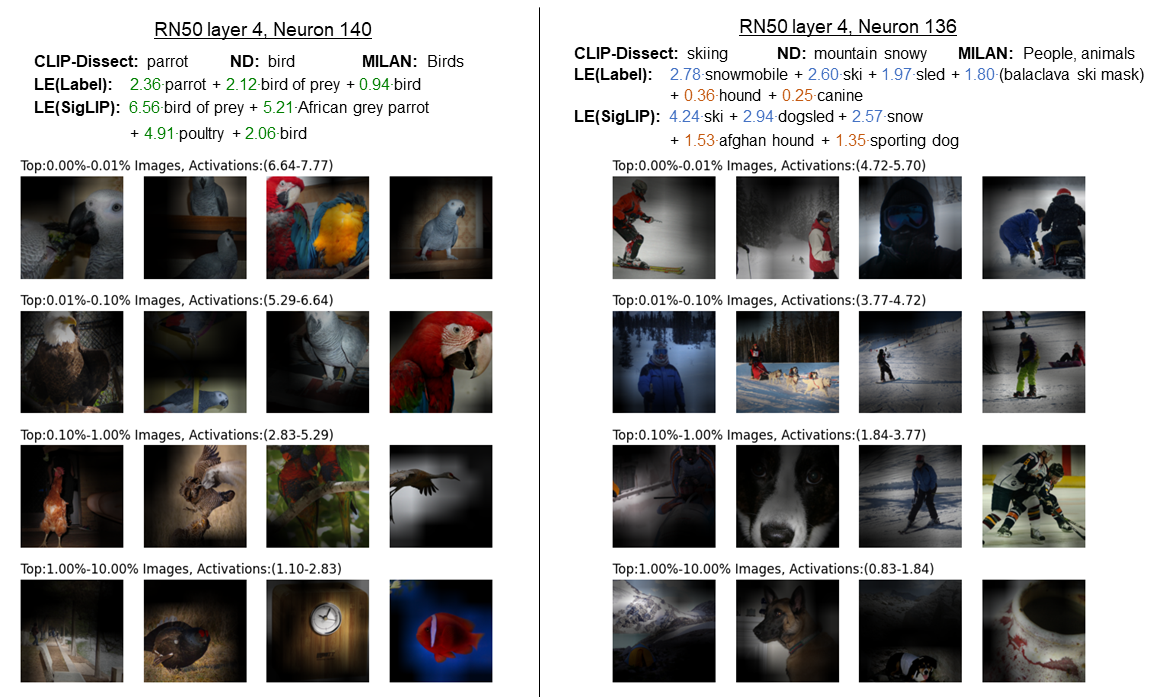}
    \caption{Descriptions and highly activating images from different ranges of example neurons. We can see Linear Explanation provides a more complete description than baselines in both cases.}
    \label{fig:example_neurons}
\end{figure*}

\paragraph{Setup}

We mostly focus our analysis on second to last layer features similar to \cite{kalibhat2023identifying} and \cite{bykov2023labeling} because they are the highest level features learned by the model, and are the features used when transfer learning from that model. Additionally, in the ResNet \cite{he2016deep} family of models, they are are CNN channels followed by a global avg pooling layer, meaning they can be meaningfully understood (and simulated) either as having 2d or scalar activations without losing any information.

We evaluate two variants of our method: \textbf{Linear Explanation (Label)}, which uses the labels in \dprobe~ to construct $P$, as well as superclass labels in the case of ImageNet and CIFAR-100. The second variant is \textbf{Linear Explanation (SigLIP)}, which uses SigLIP-ViT-L-16-384 model (different from our simulator model) to construct the concept activation matrix $P$.

\textbf{\dprobe:} As probing data we use the validation dataset of the dataset the model was trained on. We randomly split this probing data into train(70\%), validation(10\%) and test(20\%) splits. Since we simulate on the entire test split, our setting corresponds to the most challenging \textit{random-only} setting from \cite{bills2023language}.

\textbf{Concept set $\mathcal{S}$:} For \textbf{LE (Label)}, the concept set is the label names in the original dataset (+ superclass names). For \textbf{LE(SigLIP)}, we use union of the label names, the labels in Broden \cite{netdissect2017}, and a \href{https://www.desiquintans.com/nounlist}{list of 6800 English nouns}. For SigLIP we filtered these concepts to only use concepts whose average top-5 activation in \dprobe was $\geq 0.5$ to avoid using concepts not present in the data. See Appendix \ref{app:concept_set_ablation} for an ablation study using different concept sets.

\subsection{Qualitative results}

In Figure \ref{fig:area_chart}, we display example \textbf{LE(Label)} explanations, as well as an area chart visualization for a few example neurons. This choice of visualization was inspired by \cite{goh2021multimodal}, as it can visualize the neuron behavior across the entire activation range, and with our label based method we can construct them automatically. To construct this, we divided neuron activations into 8 evenly spaced buckets between 0 and the max activation of the neuron, and then plot fraction of inputs within that activation range that belong to each class/superclass. Note the later buckets have much fewer input in them. We can see that these neurons are very well explained by the Linear Explanation, and LE can help reveal polysemanticity across different activation ranges, such as Neuron 136 which activates on dog related concepts on lower activations(orange region), while its top activations are snow related concepts(blue region). Existing methods miss the dog related role and only explain the top activations, as seen in Figure \ref{fig:example_neurons}. We display explanations from ours and other methods in a more traditional way by visualizing inputs from different activation ranges in Figure \ref{fig:example_neurons}, and for many more neurons in Appendix \ref{app:additional_qualitative}.

\input{table/correlation_score_main}
\input{table/ablation_score}

\subsection{Simulation results: Correlation Scoring}

Table \ref{tab:correlation_score} shows the average correlation scores between simulated and actual neuron activations, across many different models and network architectures, including both CNNs and ViTs. We can see Linear Explanations, especially SigLIP, significantly outperform existing methods, reaching around 0.4 correlation on average, twice as high as existing methods. There is relatively large variance in how interpretable individual units are between different architectures and datasets, but performance between methods is quite consistent. The average length of our LE explanations are shown in Table \ref{tab:explanation_length}(App. \ref{app:explanation_length}). Correlation scores for other layers of a network are discussed in Appendix \ref{app:lower_results}. In Table \ref{tab:simulator_ablation}(App. \ref{app:simulator_ablation}) we perform an ablation study by using a different simulator model, showing that the trends we observe here hold across different simulators.

\subsection{Simulation results: Ablation scoring}

\begin{figure}[]
    \centering
    \includegraphics[width=0.95 \linewidth]{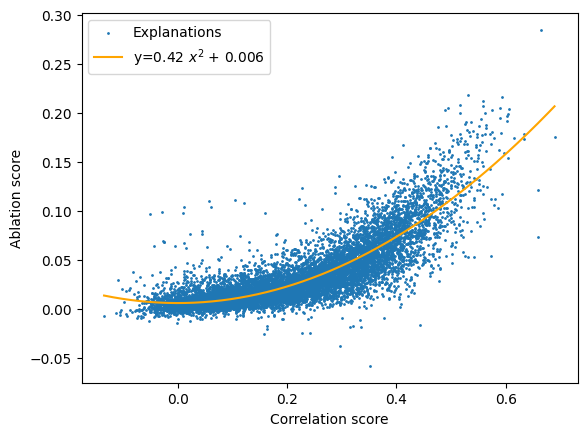}
    \caption{The relationship between correlation and ablation score of different explanations. }
    \label{fig:corr_vs_abbl}
\end{figure}

In Table \ref{tab:ablation_score}, we score the explanations using ablation scoring, reporting the average score across all neurons of layer 4 in ResNet-50. We find that existing methods perform very poorly under ablation scoring, with all methods averaging scores of at most 0.02, where simply predicting mean activation on all inputs would give a score of 0, and a perfect prediction can reach 1. Our methods improve significantly over existing methods, with Linear Explanation (SigLIP) reaching 0.0727 average ablation score, more than 3 $\times$ better than previous methods, but still score quite low overall. This highlights the need for further refinement in neuron explanation methods. This is consistent with results of \cite{bills2023language} in language models, where they found most neuron explanations scored very close to 0 under random-only ablation scoring. We also found that there is a clear quadratic relationship between the correlation and ablation score of an explanation across all explanation methods, as shown in Figure \ref{fig:corr_vs_abbl}. This shows correlation score is predictive of ablation score. 

\paragraph{Discussion.}

Overall we found that \textbf{LE(SigLIP)} outperformed \textbf{LE(Label)} by a significant margin in our evaluations. We believe this can be attributed to a few causes: First, LE(SigLIP) can use a larger concept set, and as such can detect a wider range of concepts. Second it is likely more aligned with the simulator model, i.e. thinks of concepts in a more similar manner to the simulator. In contrast, LE(Label) utilizes some concepts that are very hard for a simulator to understand based on name alone, such as \textit{leporid}, which is a superclass of rabbits. In section \ref{app:simulator_ablation} we tested using a simulator from a different model family, but SigLIP's advantage still persisted. Regardless, it may be beneficial to use LE(Label) in some cases, for example to have more transparency and consistency in the explanations.
\paragraph{Computational Efficiency.}
Both of our explanation methods take around 4 seconds per explained neuron on a machine with a single Tesla V100 GPU. In addition, \textbf{LE(SigLIP)} has a one-time cost of a single forward pass of \dprobe{} through the SigLIP model which takes around 30 minutes. This does not need to be recalculated when explaining new neurons or models with the same data.

%% file: table/correlation_score_main.tex
\begin{table*}[]
\centering

\begin{tabular}{@{}llllll@{}}
\toprule
Target model & Network Dissection & MILAN & CLIP-Dissect & LE (Label) & LE (SigLIP) \\ \midrule
ResNet-50 (ImageNet) & 0.1242 $\pm$ 0.002 & 0.0920 $\pm$ 0.002 & 0.1871 $\pm$ 0.002 & 0.2924 $\pm$ 0.002 & \textbf{0.3772 $\pm$ 0.002} \\
ResNet-18 (Places365) & 0.2038 $\pm$ 0.005 & 0.1557 $\pm$ 0.005 & 0.2208 $\pm$ 0.005 & 0.3388 $\pm$ 0.004 & \textbf{0.4372 $\pm$ 0.003} \\
VGG-16 (CIFAR-100) & n/a & n/a & 0.2298 $\pm$ 0.004 & 0.4330 $\pm$ 0.004 & \textbf{0.4970 $\pm$ 0.004} \\
ViT-B/16 (ImageNet) & n/a & n/a & 0.1722 $\pm$ 0.004 & 0.3243 $\pm$ 0.005 & \textbf{0.3489 $\pm$ 0.005} \\
ViT-L/32 (ImageNet) & n/a & n/a & 0.0549 $\pm$ 0.002 & 0.1879 $\pm$ 0.004 & \textbf{0.2182 $\pm$ 0.004} \\ \bottomrule
\end{tabular}
\caption{Average correlation scores between simulated and actual neuron activations, across all neurons in the second to last layer of the respective models. For ViT models we report the MLP neurons in the last transformer block. We do not include Network Dissection and MILAN results for the last two models, as those methods are designed for 2d activations, while the final layers of these models have effectively scalar activations.}
\label{tab:correlation_score}
\end{table*}


%% file: table/ablation_score.tex
\begin{table*}[t!]
\centering 
\begin{tabular}{@{}llllll@{}}
\toprule
Target model & \begin{tabular}[c]{@{}l@{}}Network \\ Dissection\end{tabular} & MILAN & CLIP-Dissect & LE(Label) & LE(SigLIP) \\ \midrule
ResNet-50 (ImageNet) & 0.0165 $\pm$ 0.0003 & 0.0137 $\pm$ 0.0003 & 0.0215 $\pm$ 0.0004 & 0.0433 $\pm$ 0.0007 & \textbf{0.0727 $\pm$ 0.0008} \\ \bottomrule
\end{tabular}
\caption{Average Ablation scores between simulated and actual neuron activations, average over all neurons in layer4.}
\label{tab:ablation_score}
\end{table*}

%% file: 6_related_work.tex
\section{Related Work}
\label{sec:rel_work}

\subsection{Automated Interpretability in Vision}

Several methods have been proposed to automatically explain the roles of individual neurons in neural networks. Network Dissection \cite{netdissect2017} is the first and likely most popular. They use a dataset with pixel-wise labeled concepts, and try to find concepts with high Intersection over Union (IoU) with binarized neuron activations. This method has a few downsides, such as treating neuron activations as binary and inability to detect concepts missing from their annotated dataset. Compositional Explanations \cite{mu2021compositional} extends Network Dissection to deal with polysemantic neurons, that may activate on multiple unrelated concepts, by searching for logical compositions of concepts, i.e. a neuron could activate on Cat OR Dog. This is similar to our method in that they propose a way to compose a more complex explanations out of simpler primitive concepts. However we believe linear composition is preferable over logical composition because it naturally operates on scalar activations while logical composition requires binarizing neuron activations, and linear functions are much faster to search for than logical formulas are.

Recently \cite{larosa2023fuller} proposed an Extension to Compositional Explanations, addressing the problem that Compositional Explanations only explain the very highest activations. This was motivated by findings similar to our Section \ref{sec:act_importance} showing that lower activation ranges are also important for network predictions. They propose dividing the activation range of a neuron into 5 buckets, and generating a separate compositional explanation for each. While this does provide a more complete picture of the neurons, the explanation is getting rather hard to understand, with 15 concepts per neuron that interact in complicated ways. We believe Linear Explanations are able achieve a similar level of completeness of description with much less complexity.

Other relevant methods include MILAN \cite{hernandez2022natural} which trains a neural net to describe the most highly activating inputs of a neuron, DnD \cite{bai2024describeanddissect} which utilizes pre-trained models to produce generative explanations of highest activating inputs, \cite{bau2020understanding} who propose doing Network Dissection with a Segmentation model instead of labeled data, and \cite{bykov2023labeling} who propose explaining neurons with Compositions of validation dataset labels similar to our LE(Label), but they use logical compositions instead of linear composition, and AUC to evaluate instead of ability to predict neuron activations.

Finally, recent papers CLIP-Dissect \cite{oikarinen2023clip} and FALCON \cite{kalibhat2023identifying} have proposed methods that don't require labeled concept information at all by relying on supervision from multimodal models such as CLIP \cite{radford2021learning}. This is related to how our LE(SigLIP) works, but both previous methods only focus on explaining the very highest activating inputs.

\subsection{Language Models Can Explain Neurons in LMs}

 \cite{bills2023language} proposes an Automated Interpretability approach to explain individual neurons in large language models. They propose an explanation method similar to MILAN \cite{hernandez2022natural}, and very different from ours, adapted to a language setting. They show a large model (GPT-4) the most highly activating inputs of a neuron and ask it to find what they have in common as the explanation for that neuron, which is studied in more detail by \cite{lee2023importance}.  \cite{bills2023language} also propose a new evaluation method: simulation, which evaluates explanations based on how well they can be used to predict activations on new inputs, which we improve upon and extend to vision setting in this work.

\subsection{Unreliability of explaining only top activations.}

Recent work \cite{nanfack2023adversarial, srivastava2023corrupting} have shown that Neuron explanations based on only highest activations are not robust, and can be be easily manipulated. Similarly, feature visualizations can be easily fooled as shown by \cite{geirhos2023dont}. In light of this, our hope is that more holistic explanations explaining entire activation range are less susceptible to such attacks.

\subsection{Linear Probes}

Linear Probing is a common method for finding explainable neurons in language models \cite{alain2018understanding, sajjad2022neuronlevel, gurnee2023finding, fong2018net2vec}. In linear probing, a (sparse) linear model is trained to predict a concept based on neuron activations, with the goal to understand if/where the network represents that concept. Our method is effectively the \textit{inverse} of linear probes, where we instead learn a sparse linear function of concepts to predict a neuron activation. 

%% file: 7_conclusion.tex
\section{Conclusion}

We have shown that only describing highly activating inputs as done by previous work is not sufficient to understanding a neuron, and proposed a new solution called \textbf{Linear Explanations}. Our method explains neurons as linear combination of concepts that produces highly accurate and complete neuron descriptions that are still easy to comprehend. Additionally, we developed a new way to more rigorously evaluate neuron explanations in vision models via simulation.

%% file: A_Appendix.tex
\newpage
\appendix
\onecolumn

\section{Appendix: Additional Information}

\subsection{Limitations}

\label{sec:limitations}

One limitation of our current method is that it does not take into account the location specific activations. This does not change our results on second-to-last layer neurons which is our main focus, but hinders our ability to explain and simulate lower layers well. However this not a fundamental feature of the method, rather an effect of the tools and data we use. Concept-activation matrix $P$ doesn't have to be at the level of images. We can similarly instead run the same method with pixel/superpixel level supervision using for example Broden \cite{netdissect2017} labels, or an open-vocabulary semantic segmentation model such as \cite{yu2023convolutions}. This will however increase the computational cost as we will have to deal with much larger $P$, and is a limitation we aim to address in the future. 

Another limitation is that the simulation scores we achieve, while better than existing methods are still quite low, especially ablation. While this is in line with previous work \cite{bills2023language}, it is worth discussing. This could be caused by 3 things: 1. Neurons are inherently not interpretable, 2. Our explanations are not good enough or 3. Our simulator is not good enough. While significant gains can definitely be made improving the simulator and explanations (and they can easily be replaced in our pipeline by more powerful models in the future), we believe a large portion of the challenge is caused by the neurons inherently not having a simple explanation. However, recent work in language models \cite{bricken2023monosemanticity, cunningham2023sparse} shows promise that we can generate more interpretable units of inspection via methods like sparse autoencoders, to which our method can be applied out-of-the-box. 

\subsection{Additional Related Work: Input Importance}

A lot of classical explainable AI methods are focused on a different problem from ours, specifically that of \textbf{Input Importance}, where the goal is to answer the question: \textit{Which parts of input $x$ are the most important in the model making prediction $f(x)$?}. 
In contrast, our work is a neuron explanation method, where the goal of our work is to answer the question: \textit{Given a neural network model $f$, what are the functionalities of each individual neuron?}. 

For example, the LIME paper \cite{ribeiro2016should} is a popular Input Importance work using linear methods for that task. We will use it to illustrate the major difference between the input importance methods and our neuron explanation method. 

3 main differences between our method and LIME \cite{ribeiro2016should}:

\begin{enumerate}
    \item \textbf{Goal}: LIME aims to explain the final prediction of a model with respect to an input, while we explain a single hidden layer neuron.
    \item \textbf{Scope}: We produce global explanations, while LIME learns a local explanation that only works close to a specific input.
    \item \textbf{Input:} Lime learns a linear model in terms of input features (or groups of input features), while our method creates an explanation based on interpretable concepts.
\end{enumerate}

\subsection{Greedy search algorithm}

Algorithm \ref{alg:greedy_search} shows the details of our greedy search procedure discussed in Section \ref{sec:method}. For the tolerance $\epsilon$, we only add another concept if it improves the correlation(on validation set) between predicted and actual neuron values by more than $\epsilon$. In our experiments, $(v, r, \epsilon)$ are set to be 10, 10, 0.02 respectively.

\label{app:greedy_search_alg}
\begin{algorithm}[h]
    \caption{Greedy search. Python pseudo-code.}
    \label{alg:greedy_search}
\begin{algorithmic}
    \STATE SC = \{\} \#selected concepts
    \STATE BC = \{\} \#bad concepts
    \STATE $\rho^* = 0$ \#best correlation
    \WHILE{$|\text{SC}| < v$}
    \STATE CC = \{\}
    \WHILE{$|CC| < r$}
        \STATE CC.add($\text{argmax}_{i \notin CC \cup SC \cup BC} (w_{k, i})$)
    \ENDWHILE
    \STATE $\rho' = \rho^*$ \#current best corr
    \FOR{$j$ in CC}
        \STATE model = train\_model($P_{SC \cup j}^{\text{train}}$, $q_k^{\text{train}}$)
        \STATE $\rho$ = get\_correlation(model, $q_k^{\text{val}}$)
        \IF{$\rho < \rho^*+\epsilon$}
            \STATE BC.add($j$) \#concept is bad
        \ELSIF{$\rho > \rho'$}
            \STATE $\rho' = \rho$
            \STATE best\_c = j
        \ENDIF    
    \ENDFOR
    \IF{$\rho'< \rho^* + \epsilon$}
        \STATE \textbf{break} \#early stop if no improvement
    \ELSE
        \STATE SC.add(best\_c)
        \STATE $\rho^* = \rho'$
    \ENDIF
    \ENDWHILE
    \STATE return SC
\end{algorithmic}
\end{algorithm}

\subsection{Explanation length}
\label{app:explanation_length}

\input{table/explanation_length}

Table \ref{tab:explanation_length} shows the average explanation lengths of our methods for second to last layers of different models. We can see that explanation lengths vary quite a bit, with in particular ViT neurons having short explanations. We think this is likely caused by the fact that the MLP neurons we studied in ViT seem to be more bimodal, either being very uninterpretable or rather monosemantic and interpretable, both of which result in short explanations in our case. If more similar explanation lengths are desired, this can be achieved by tuning the tolerance parameter of the greedy search for each model.

\newpage

%% file: table/explanation_length.tex
\begin{table}[h!]
\centering
\begin{tabular}{@{}lll@{}}
\toprule
Target model & LE (Label) & LE (SigLIP) \\ \midrule
ResNet-50 (ImageNet) & 4.37 & 4.68 \\
ResNet-18 (Places365) & 4.70 & 4.25 \\
VGG-16 (CIFAR-100) & 7.60 & 6.08 \\
ViT-B/16 (ImageNet) & 1.97 & 1.82 \\
ViT-L/32 (ImageNet) & 1.53 & 1.51 \\ \bottomrule
\end{tabular}
\caption{Average Explanation lenghts of our method for second to last layer neurons of different models discussed in Table \ref{tab:correlation_score}.}
\label{tab:explanation_length}
\end{table}

%% file: B_Appendix_Ablations.tex
\clearpage
\newpage

\section{Appendix: Ablations}

\subsection{Different simulator model.}
\label{app:simulator_ablation}
Since both our simulator and SigLIP explainer model use a similar (but different) model from the SigLIP family, it is natural to ask whether the good performance is caused by this similarity of models. To address this, we redo our simulation with correlation scoring experiment (Table \ref{tab:correlation_score}) using a simulator model from original CLIP \cite{radford2021learning} family, namely CLIP-ViT-L-14-336. Results are shown in Table \ref{tab:simulator_ablation}. We can see the that all scores are lower than with our original simulator model (SigLIP-SO400M-14-386), likely because this CLIP model is not as powerful and it is not designed for multilabel classification required from a simulator. Our Linear Explanation (SigLIP)'s performance is reduced the most by switching the simulator model, indicating that using a related simulator model does boost it's performance, but it is still the best method with a different simulator.

\input{table/simulator_ablation}

\subsection{Different concept sets}
\label{app:concept_set_ablation}

In table \ref{tab:concept_set_ablation}, we tested how the choice of concept set affects the results of our LE(SigLIP). We compared our original (class labels + Broden labels + common nouns) against two alternatives: 20k, a set of 20,000 most common English words from \cite{oikarinen2023clip}, as well as only using the ImageNet class labels. We can see our concept set noticeably outperforms both alternative choices. Interestingly it even outperforms the larger 20k concept set, perhaps because it is more noisy and lacking some more precise terms(class names) useful in describing ImageNet images.

\input{table/ablation_concept_set}

\subsection{Different explainer model}

We also conducted a study testing the importance of the choice of explainer model for the performance of LE(SigLIP). In particular, we replaced our original explainer SigLIP-ViT-L-16-384 with CLIP ViT-L-14-336 from the original CLIP paper \cite{radford2021learning}. From the results in Table \ref{tab:ablation_explainer} we can see a stronger explainer model makes a big difference, but even when using a weaker CLIP model our method outperforms other methods.

\input{table/ablation_explainer_model}

\subsection{No greedy search}
\label{app:gs_ablation}

To assess how important our greedy search procedure described in section \ref{sec:greedy_search} is for explanation quality, we evaluated our method without the greedy search procedure by simply taking the top-k concepts of our relatively sparse $w_k$ (with weights retrained with top-k concepts only) as the final explanation for the neuron. Simulation results are shown in Table \ref{tab:gs_ablation}. Overall the results are mixed, showing we can reach roughly similar quality explanations without the greedy search, but we still find using it preferable as greedy search can dynamically determine the desired explanation length for each neuron, giving longer descriptions to more complex neurons and simple explanations to monosemantic neurons.

\input{table/greedy_search_ablation}


\subsection{Scaling for Simulation with Ablation Scoring}
\label{app:ablation_scaling}

In section \ref{sec:simulation} we define our ablation scoring function $\alpha_{init}(k, \mathcal{D}, E, c, d)$. This requires finding the optimal scaling parameters $c^*$ and $d^*$. For our main results we used the optimization method for finding these parameters:

\textbf{Optim:}
\begin{equation}
    c^*, d^* = \text{arg}\max_{c,d} \alpha_{init}(k, \mathcal{D}_{\text{probe}}^{\text{val}}, E, c, d) 
\end{equation}

In contrast, \cite{bills2023language} use a closed form solution intended to maximize the explained variance to select these parameters. In their method (which we will call \textbf{Norm}):

\textbf{Norm:}

Let
\begin{equation}
    \rho_{\text{val}}(k, E) = \sum_{x \in \mathcal{D}_{\text{probe}}^{\text{val}}} \frac{\hat{s}(x, E) \cdot \hat{g}(A_k(x))}{|\mathcal{D}_{\text{probe}}^{\text{val}}|}
\end{equation}

where $\hat{s}$ and $\hat{g}(A_k(x))$ are normalized to have mean 0 and standard deviation 1 on the validation set. Then: 

\begin{equation}
    c^* = \rho_{\text{val}}(k, E) \cdot \frac{\sigma(\mathcal{D}_{\text{probe}}^{\text{val}}, g(A_k(\cdot)))}{\sigma(\mathcal{D}_{\text{probe}}^{\text{val}}, s(\cdot, E))}
\end{equation}

\begin{equation}    
    d^* =  - c^* \cdot \mu(\mathcal{D}_{\text{probe}}^{\text{val}}, s(\cdot, E)) + \mu(\mathcal{D}_{\text{probe}}^{\text{val}}, g(A_k(\cdot)))
\end{equation}

where:
\begin{equation}
    \mu(\mathcal{D}, f) = \frac{1}{|\mathcal{D}|} \sum_{x_i \in \mathcal{D}} f(x_i)
\end{equation}

\begin{equation}
    \sigma(\mathcal{D}, f) = \sqrt{\frac{\sum_{x_i \in \mathcal{D}} (f(x_i) - \mu(\mathcal{D}, f))^2}{|\mathcal{D}|}}
\end{equation}

This is theoretically justified as maximizing the explained variance of the neuron. However, our goal is not to maximize explained variance, which leads us to believe direct optimization might be more effective (\textbf{Optim}). In table \ref{tab:ablation_scaling} we compare the two methods when evaluating explanations for layer4 neurons of ResNet-50(ImageNet). We can see the optimization method results in noticeably (20-50\%) larger ablation scores overall, but doesn't affect the order between explanation methods.

However, optim requires several (100 in our experiments) backwards passes over the validation data, and as such is much more computationally costly. For second to last layer neurons this can be done very efficiently with only a few seconds per neuron, but for evaluating earlier layers the \textbf{Norm} strategy is likely better.

\input{table/ablation_scaling}

\newpage



%% file: table/simulator_ablation.tex
\begin{table}[h!]
\begin{tabular}{llllll}
\hline
Target model & Network Dissection & MILAN & CLIP-Dissect & \begin{tabular}[c]{@{}l@{}}Linear Explanation\\ (Label)\end{tabular} & \begin{tabular}[c]{@{}l@{}}Linear Explanation\\ (SigLIP)\end{tabular} \\ \hline
ResNet-50 (ImageNet) & 0.1003 & 0.0789 & 0.1725 & 0.2545 & \textbf{0.2961} \\
ResNet-18 (Places365) & 0.1626 & 0.1279 & 0.1951 & 0.2825 & \textbf{0.3154} \\ \hline
\end{tabular}
\caption{Correlation scores of different explanations in layer4 of the models, using CLIP ViT-L-14-336 as the simulator model. Results are similar to Table \ref{tab:correlation_score}, but all methods score lower, probably due to weaker simulator model.}
\label{tab:simulator_ablation}
\end{table}

%% file: table/ablation_concept_set.tex
\begin{table}[h!]
\centering
\begin{tabular}{@{}ll@{}}
\toprule
Concept set & Correlation score \\ \midrule
Original (8438) & \textbf{0.3772} \\
20k (20,000) & 0.3561 \\
ImageNet labels (1000) & 0.3378 \\ \bottomrule
\end{tabular}
\caption{Average simulation correlation scores for final layer neurons of ResNet-50(ImageNet) layer4 of LE(SigLIP) with different choices of concept set. We can see our concept set performs the best. Number in brackets represents the size of the concept set}
\label{tab:concept_set_ablation}
\end{table}

%% file: table/ablation_explainer_model.tex
\begin{table}[h!]
\centering
\begin{tabular}{@{}lll@{}}
\toprule
Explainer model & Correlation score &  \\
\midrule
Original (SigLIP-ViT-L-16-384) & \textbf{0.3772} &  \\
CLIP ViT-L-14-336 & 0.3243 & \\
\bottomrule
\end{tabular}
\caption{Comparison of different explainer models for LE(SigLIP) on laeyr4 neurons of ResNet-50(ImageNet).}
\label{tab:ablation_explainer}
\end{table}

%% file: table/greedy_search_ablation.tex
\begin{table}[h!]
\centering
\begin{tabular}{@{}lll@{}}
\toprule
Method & LE(label) & LE(siglip) \\ \midrule
Original & 0.2924 (4.37) & \textbf{0.3772 (4.68)} \\
No greedy search (top-5) & \textbf{0.3136 (5.00)} & 0.3767 (5.00) \\
No greedy search (top-4) & 0.2962 (4.00) & 0.3548 (4.00) \\ \bottomrule
\end{tabular}
\caption{Average correlation score of different explanations for neurons in layer4 of ResNet-50(ImageNet). Number in brackets is the average description length.}
\label{tab:gs_ablation}
\end{table}

%% file: table/ablation_scaling.tex
\begin{table}[h!]
\centering
\begin{tabular}{llllll}
\hline
Scaling Method & \begin{tabular}[c]{@{}l@{}}Network \\ Dissection\end{tabular} & MILAN & CLIP-Dissect & LE(Label) & LE(SigLIP) \\ \hline
Norm & 0.0118 $\pm$ 0.0004 & 0.0092 $\pm$ 0.0004 & 0.0187 $\pm$ 0.0005 & 0.0334 $\pm$ 0.0007 & \textbf{0.0602 $\pm$ 0.0009} \\
Optim & 0.0165 $\pm$ 0.0003 & 0.0137 $\pm$ 0.0003 & 0.0215 $\pm$ 0.0004 & 0.0433 $\pm$ 0.0007 & \textbf{0.0727 $\pm$ 0.0008} \\ \hline
\end{tabular}
\caption{Testing the scaling function for ablation scoring with simulation. Average ablation score across all neurons in layer4 of ResNet-50(ImageNet). We can see optimizing the scaling parameters $c$, $d$ results in a 20-50\% increase in ablation scores, while not affecting the ordering between methods.}
\label{tab:ablation_scaling}
\end{table}

%% file: C_Appendix_more_results.tex
\section{Appendix: Additional Results}

\subsection{Additional layers.}

\label{app:lower_results}

\paragraph{ResNet lower layer neurons.} Table \ref{tab:correlation_layer} and \ref{tab:rn18_layers} show the correlation scores of different methods across different layers of ResNet-50(ImageNet) and ResNet-18(Places365). We can see performance significantly degrades on the lower layers, which we think is partially caused by neuron functions being harder to describe, but also because our simulator cannot simulate pixel level activations, which becomes more impactful when describing lower layer neurons. However we can still see Linear Explanation, especially with SigLIP continues to outperform all existing methods. Finally in Table \ref{tab:vgg16_layers} we report the results on two different layers on VGG-16.

\input{table/correlation_by_layer}
\input{table/rn18_layers}
\input{table/vgg16_layers}

\paragraph{Different layers of ViT}

In table \ref{tab:vit_layers}, we study the interpretability of different layers in ViT models.
We can see that the residual stream neurons are in general noticeably less interpretable than the MLP neurons. This is likely caused by the fact that the residual stream does not (mostly) have a "priviliged basis", which means individual neurons are not more interpretable than random directions in the activation space. In other layers/architectures such a priviliged basis is created by axis-aligned activation functions such as ReLU. See \cite{elhage2023priviliged} for more discussion on priviliged bases. Because of this, we focus on analyzing the MLP neurons in our work. This is in line with work investigating individual neurons in transformer language models, which typically focus on these MLP neurons. Interestingly, we found that some of the neurons in the MLP layers were extremely interpretable with the highest correlation scores $>0.9$, which we did not see in CNN models. On the other hand, the MLP layer also had several lowly activating/dead neurons that were not particularly interpretable, bringing down the average.

When evaluating the final layer neurons of ViT (both residual stream and MLP), we only recorded the activations of the CLS token, as this is the only part passed on to the classification head, and the other activations do not matter. For earlier ViT layers, we took average across the CLS token and all spatial activations, though further research is needed to better know how important the CLS token is compared to spatial activations at different layers of the network.

\input{table/vit_layers}

\subsection{Dead neurons in ViT}

When investigating the neurons in Vision Transformer, we came across a curious phenomenon: several neurons of the MLP layers are completely dead, i.e. don't activate on any inputs. Such neurons are not meaningful to explain, and seem to waste model capacity. This was especially noticeable in our Experiments with ViT-L/16, where every single neuron in the last transformer block is dead. We tested this further, and found that the last (23rd) transformer layer does not do anything, and can be deleted from the model without affecting classification accuracy at all. Because of this weirdness, we did not include ViT-L/16 in our main results and instead used ViT-L/32.

In Table \ref{tab:dead_neurons} we quantify the number of dead neurons in the second to last layer of different models (same layers as in Table \ref{tab:correlation_score}). We defined a neuron as dead if it's maximum activation was $<0.01$ across the probing data. We can see other ViT models suffer from some dead neurons but not very many, with ViT-L/16 being an outlier. The CNN based models have no dead neurons. 

\input{table/dead_neurons}

In Table \ref{tab:correlation_no_dead}, we report the average correlation scores on "active" neurons only, i.e.we  don't count the dead neurons for models with some dead neurons. Overall this does not make a big change as the fraction of dead neurons was quite small, but most scores improve by around 5\% over Table \ref{tab:correlation_score}, which is roughly the fraction of dead neurons.

\input{table/correlation_without_dead}

\newpage

\subsection{Additional Qualitative Results}

\begin{figure}[h]
    \centering
    \includegraphics[width=0.95\linewidth]{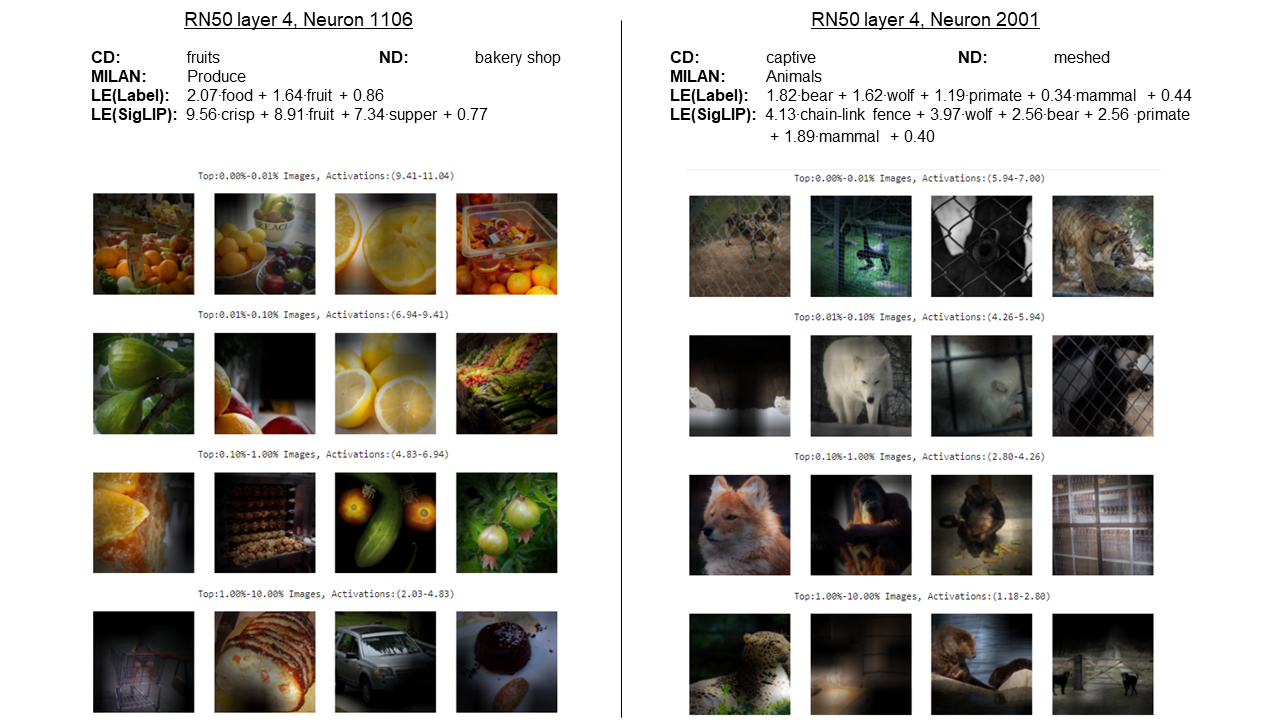}
    \caption{Example interpretable neurons.}
    \label{fig:nice_example_2}
\end{figure}

\begin{figure}[h]
    \centering
    \includegraphics[width=0.95\linewidth]{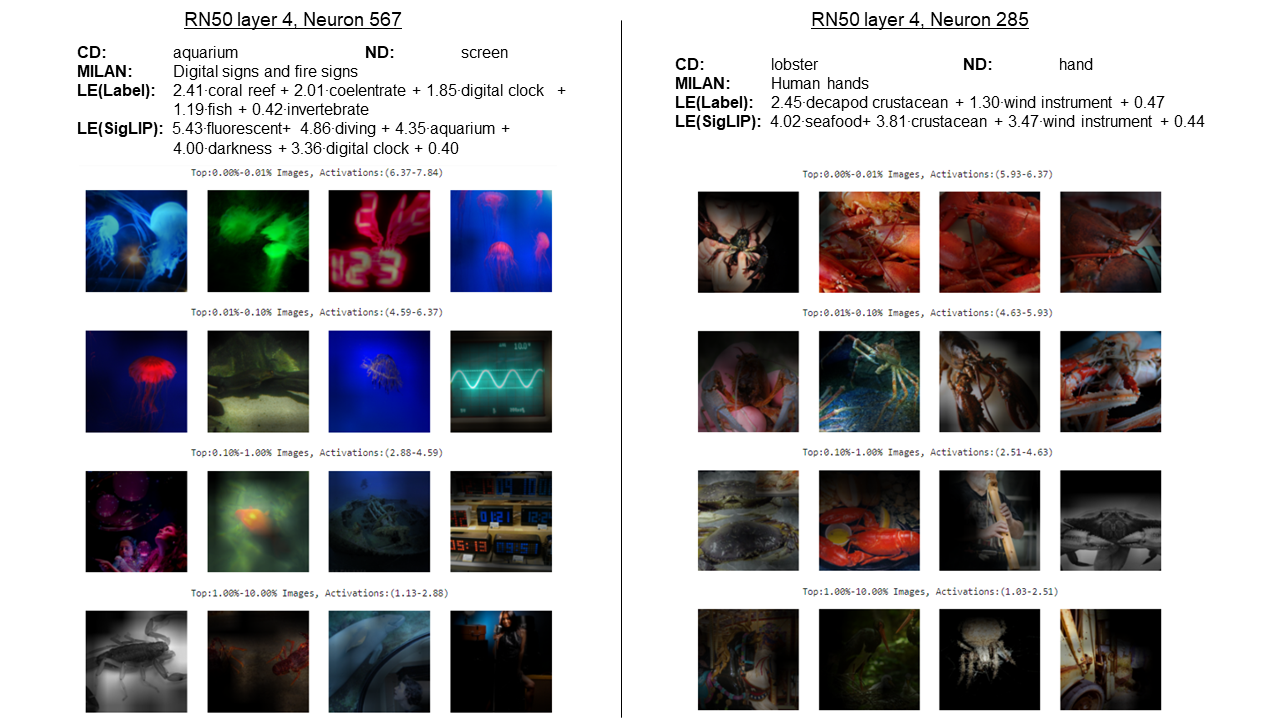}
    \caption{Example interpretable neurons.}
    \label{fig:nice_example_3}
\end{figure}

\begin{figure}
    \centering
    \includegraphics[width=0.95\linewidth]{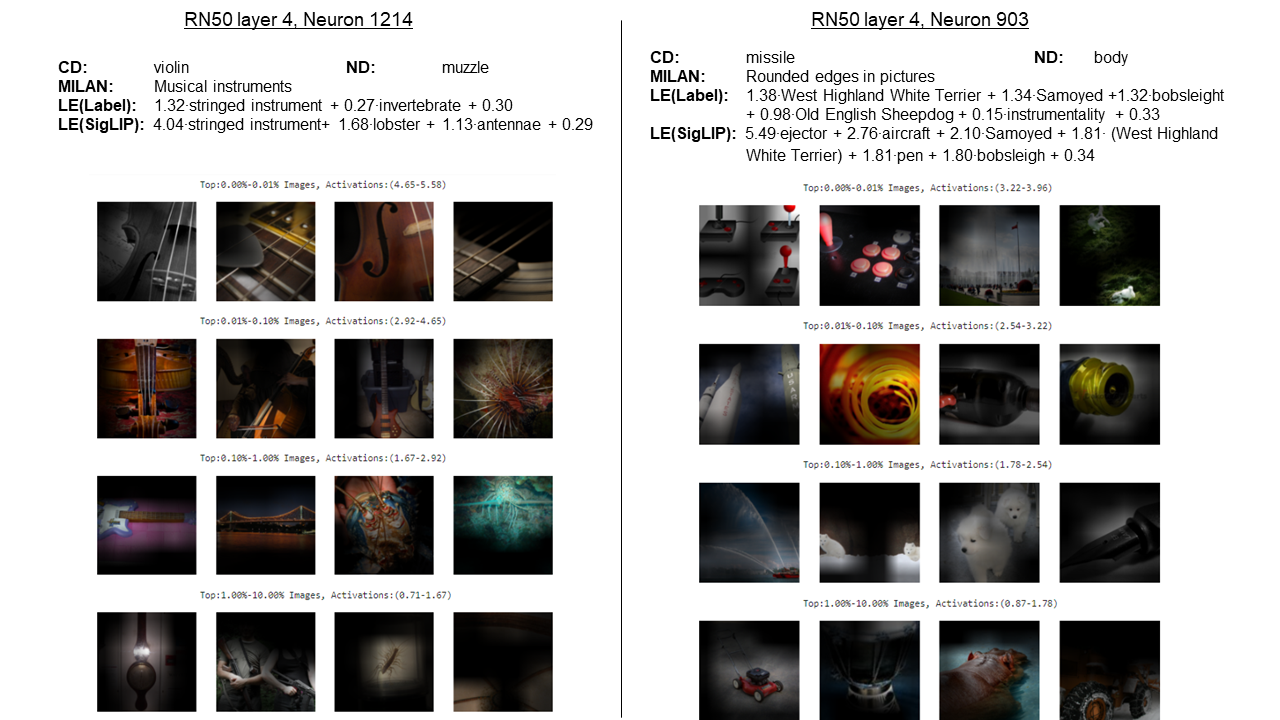}
    \caption{Randomly selected neurons.}
    \label{fig:random_example_1}
\end{figure}

\begin{figure}
    \centering
    \includegraphics[width=0.95\linewidth]{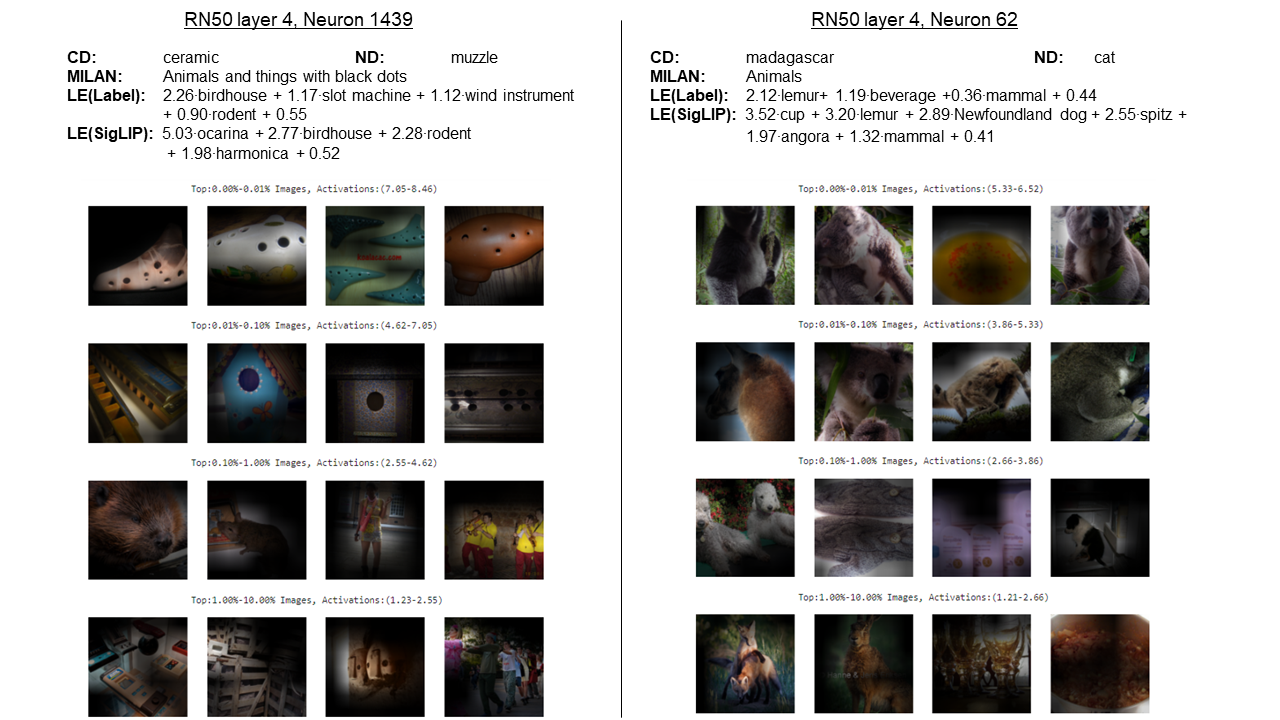}
    \caption{Randomly selected neurons.}
    \label{fig:random_example_2}
\end{figure}

\begin{figure}
    \centering
    \includegraphics[width=0.95\linewidth]{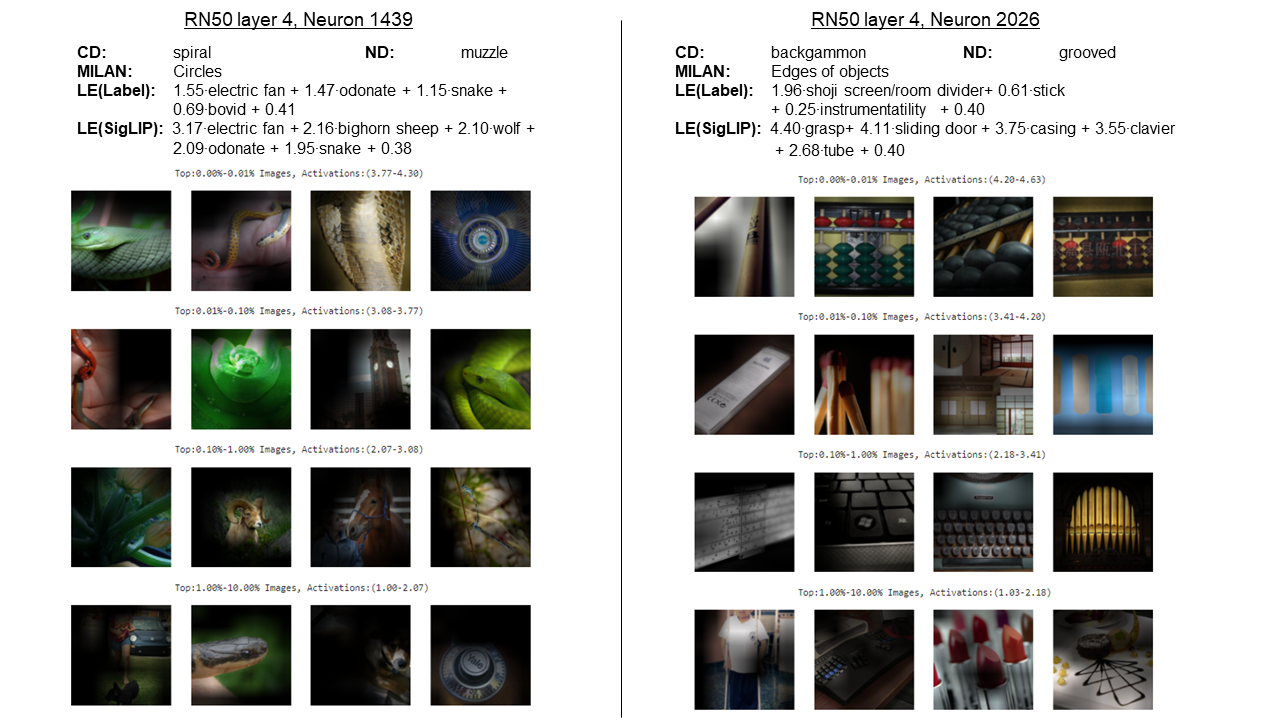}
    \caption{Randomly selected neurons.}
    \label{fig:random_example_3}
\end{figure}

In Figures  \ref{fig:nice_example_2}, \ref{fig:nice_example_3}, \ref{fig:random_example_1}, \ref{fig:random_example_2}, \ref{fig:random_example_3} we display our explanations as well as previous work for several neurons from layer4 of ResNet-50, and show that our explanations generally capture the neurons behavior well, even if it doesn't have a simple function. In addition, we notice that the LE(Label) and LE(SigLIP) methods typically produce similar concepts, but SigLIP version often returns higher weights for these. This is likely due to differences in concept activation matrices $P$ used by the two models, and highlights the need to calibrate the magnitude of predictions when simulating to align these with the simulator.

\begin{figure}
    \centering
    \includegraphics[width=0.95\linewidth]{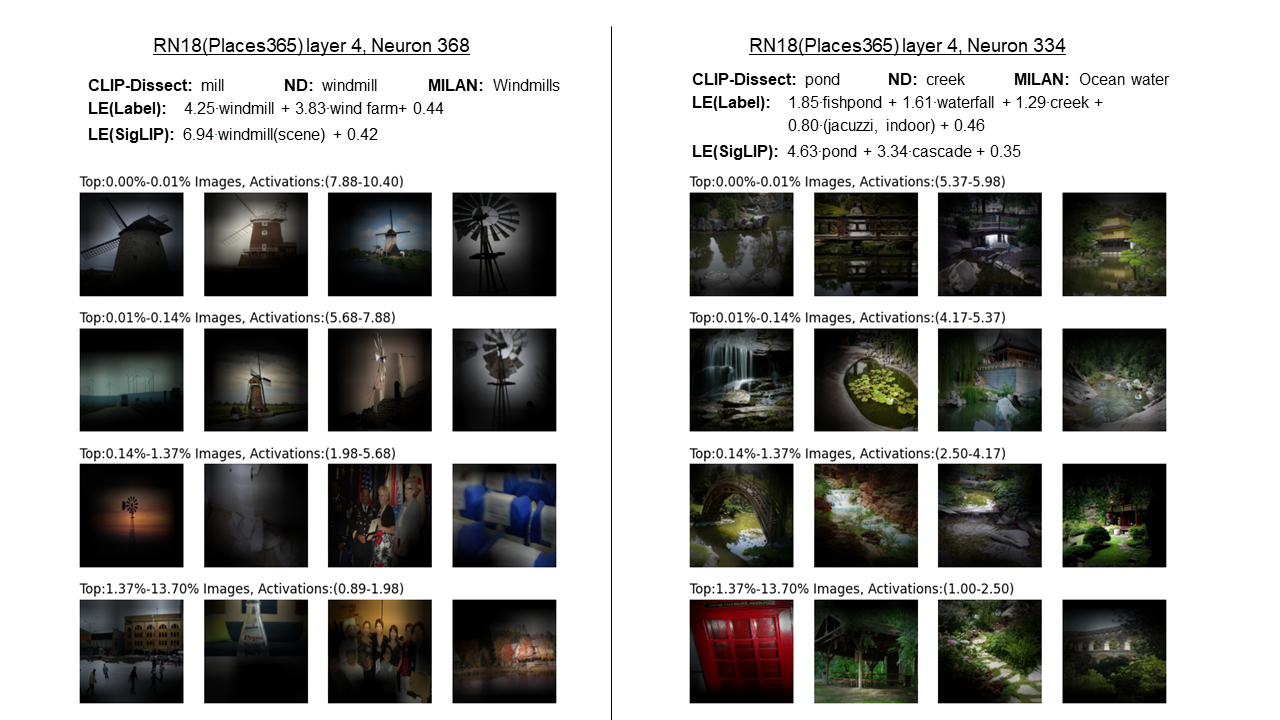}
    \caption{Example interpretable neurons of ResNet-18(Places365).}
    \label{fig:places_example1}
\end{figure}

\begin{figure}
    \centering
    \includegraphics[width=0.95\linewidth]{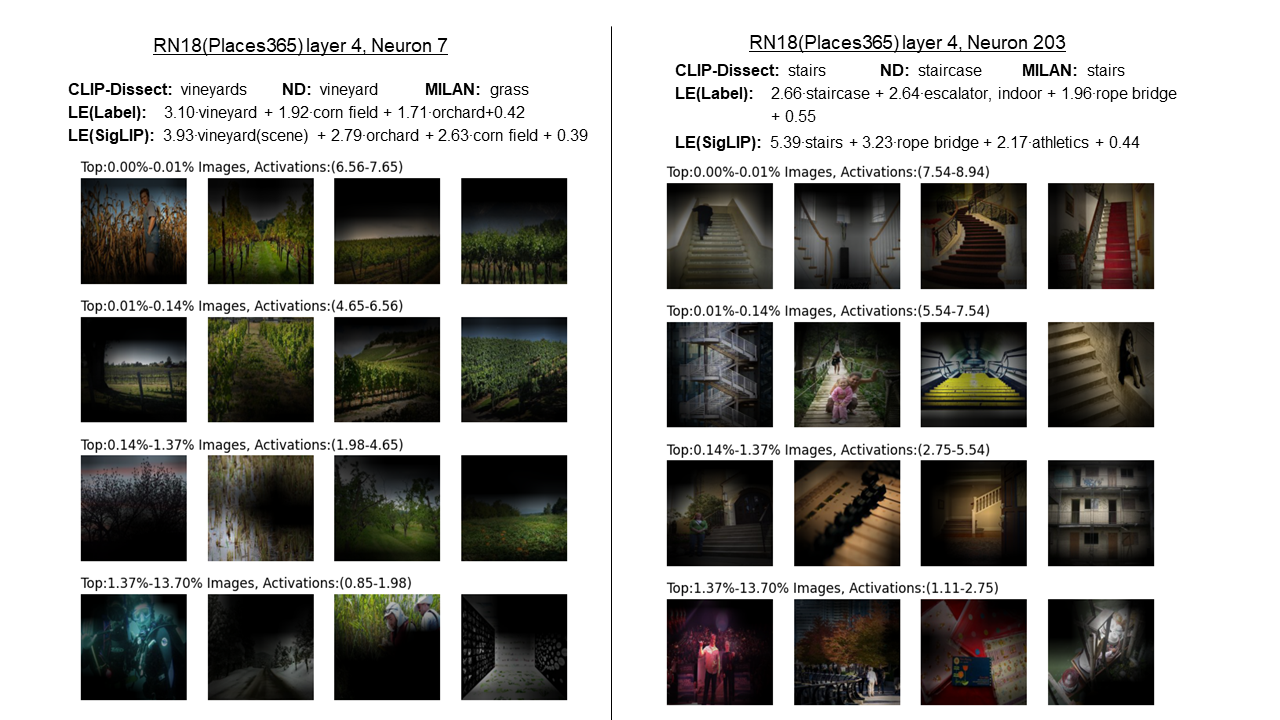}
    \caption{Example interpretable neurons of ResNet-18(Places365).}
    \label{fig:places_example2}
\end{figure}

\begin{figure}
    \centering
    \includegraphics[width=0.95\linewidth]{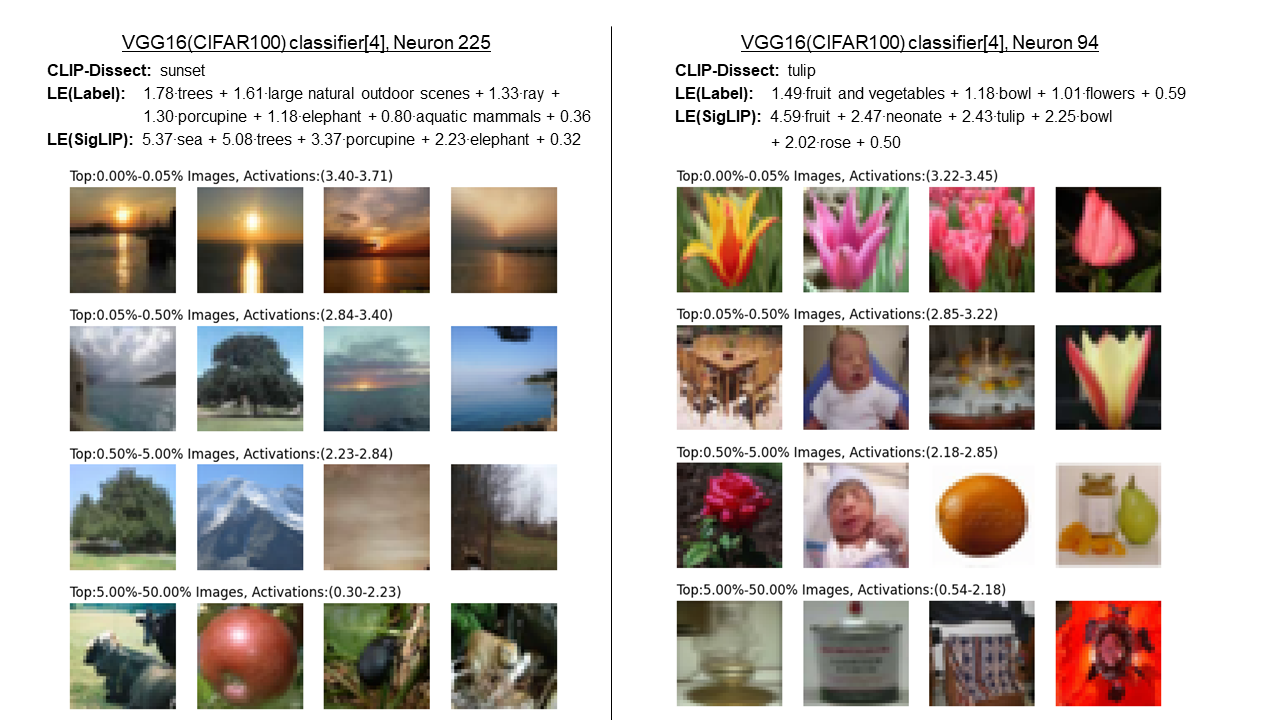}
    \caption{Example interpretable neurons of VGG-16(CIFAR-100).}
    \label{fig:vgg_example}
\end{figure}

\begin{figure}
    \centering
    \includegraphics[width=0.95\linewidth]{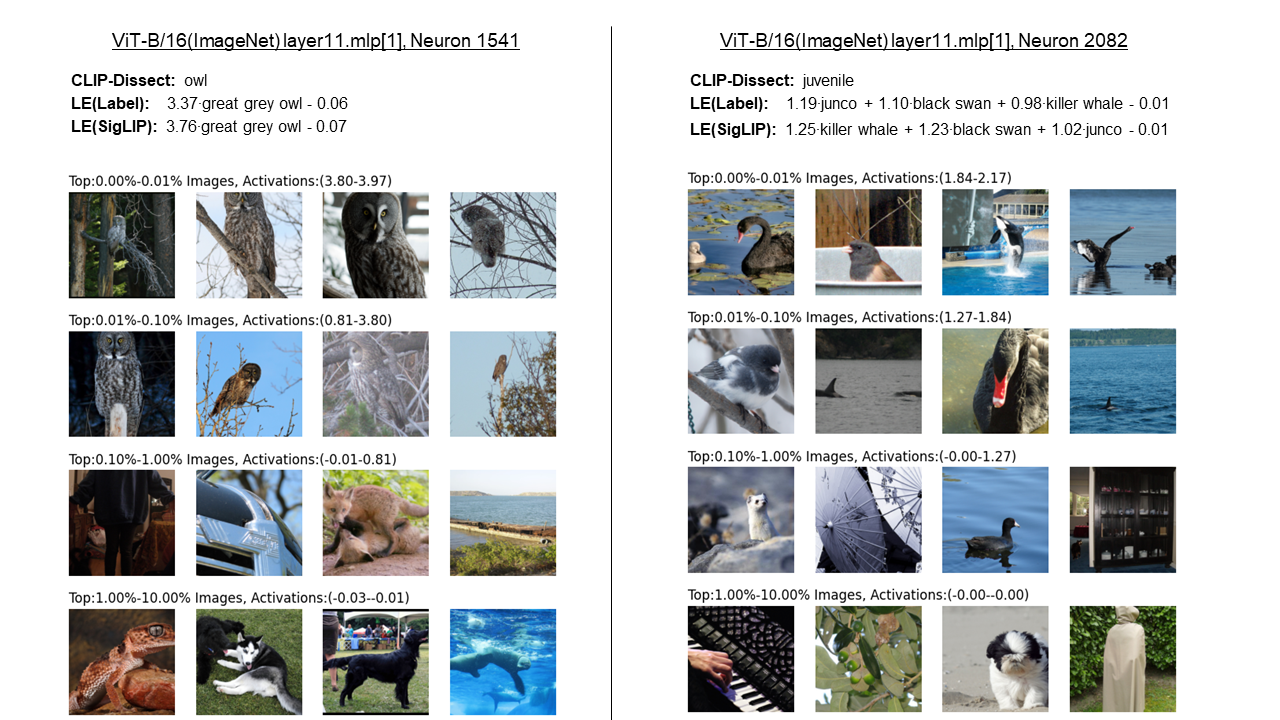}
    \caption{Example interpretable neurons of ViT-B/16(ImageNet).}
    \label{fig:vit_b_16_example}
\end{figure}

\begin{figure}
    \centering
    \includegraphics[width=0.95\linewidth]{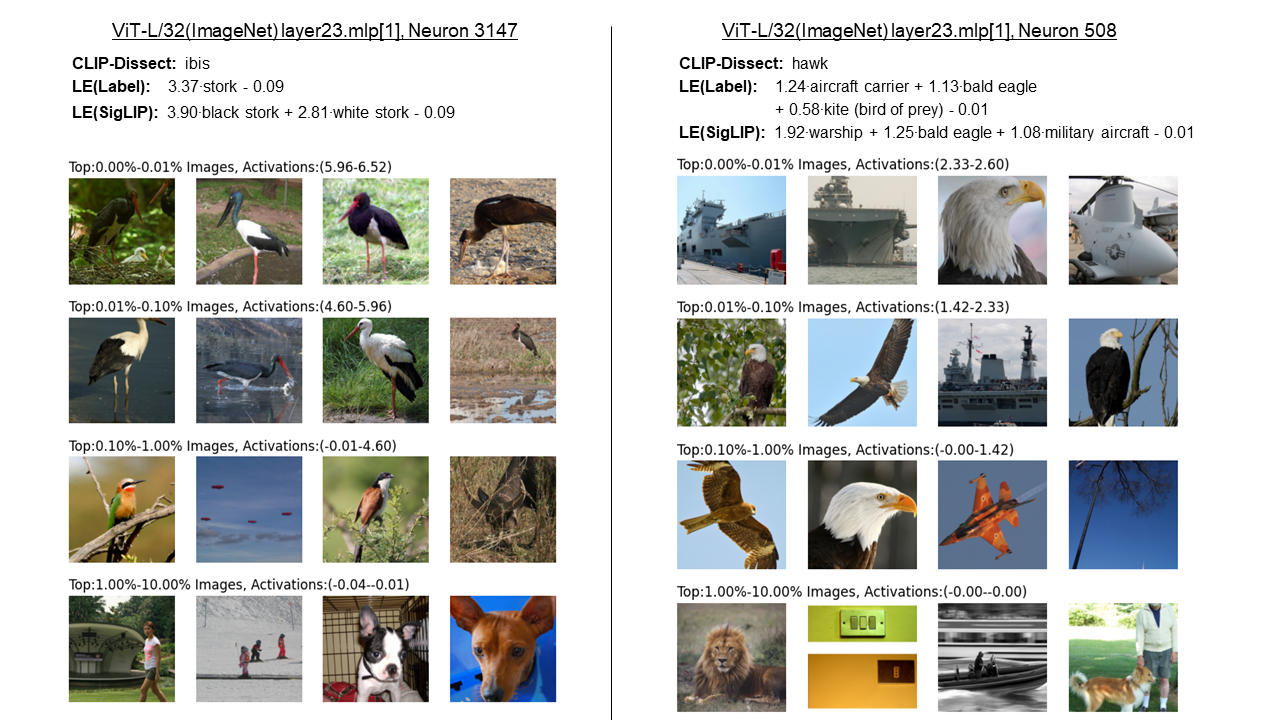}
    \caption{Example interpretable neurons of ViT-L/32.}
    \label{fig:vit_l_32_example}
\end{figure}

In Figures \ref{fig:places_example1}, \ref{fig:places_example2}, \ref{fig:vgg_example}, \ref{fig:vit_b_16_example}, \ref{fig:vit_l_32_example} we display similar figures for other models. Some interesting things we can see are that in the case on extremely monosemantic neurons, such as ViT-B/16 neuron 1541 in Figure \ref{fig:vit_b_16_example}, our method correctly returns only one concept explanation for that neuron, and that concept is more accurate than that of CLIP-Dissect. We also found a rather interesting neuron activating on both military ships/aircraft as well as bald eagles as shown in Figure \ref{fig:vit_l_32_example}, which our methods described very accurately.

\label{app:additional_qualitative}

%% file: table/correlation_by_layer.tex
\begin{table*}[h!]
\centering
\begin{tabular}{@{}llllll@{}}
\toprule
Target layer & Network Dissection & MILAN & CLIP-Dissect & \begin{tabular}[c]{@{}l@{}}Linear Explanation \\ (Label)\end{tabular} & \begin{tabular}[c]{@{}l@{}}Linear Explanation \\ (SigLIP)\end{tabular} \\ \midrule
layer1 & 0.0313 & 0.0430 & 0.0670 & 0.1008 (1.98) & \textbf{0.2313 (3.32)} \\
layer2 & 0.0324 & 0.0242 & 0.0527 & 0.1037 (1.91) & \textbf{0.2180 (3.22)} \\
layer3 & 0.0794 & 0.0773 & 0.1045 & 0.1308 (2.32) & \textbf{0.2652 (3.69)} \\
layer4 & 0.1231 & 0.0920 & 0.1871 & 0.2924 (4.37) & \textbf{0.3772 (4.68)} \\ \bottomrule
\end{tabular}
\caption{Average correlation scores of explanations by different methods on different layers of ResNet-50 trained on ImageNet. We can see correlation scores noticeably decrease on lower layers, but Linear Explanations perform the best on all layers. The number in brackets represents average number of concepts per explanation.}
\label{tab:correlation_layer}
\end{table*}

%% file: table/rn18_layers.tex
\begin{table*}[h!]
\centering
\begin{tabular}{@{}llllll@{}}
\toprule
Target layer & Network Dissection & MILAN & CLIP-Dissect & \begin{tabular}[c]{@{}l@{}}Linear Explanation \\ (Label)\end{tabular} & \begin{tabular}[c]{@{}l@{}}Linear Explanation \\ (SigLIP)\end{tabular} \\ \midrule
layer1 & 0.0223 & 0.0344 & 0.0603 & 0.1384 (2.42) & \textbf{0.2574 (3.66)} \\
layer2 & 0.0261 & 0.0154 & 0.0419 & 0.1351 (2.41) & \textbf{0.2803 (3.84)} \\
layer3 & 0.0727 & 0.0768 & 0.0704 & 0.1593 (2.73) & \textbf{0.2842 (3.71)} \\
layer4 & 0.2038 & 0.1557 & 0.2208 & 0.3388 (4.70) & \textbf{0.4372 (4.25)} \\ \bottomrule
\end{tabular}
\caption{Average correlation scores of explanations by different methods on different layers of ResNet-18 trained on Places365. The number in brackets represents average number of concepts per explanation.}
\label{tab:rn18_layers}
\end{table*}

%% file: table/vgg16_layers.tex
\begin{table}[h!]
\centering
\begin{tabular}{@{}lllll@{}}
\toprule
Model & Target layer & CLIP-Dissect & LE(label) & LE(SigLIP) \\ \midrule
VGG16 (CIFAR100) & \begin{tabular}[c]{@{}l@{}}classifier{[}4{]} \\ (second to last, fc)\end{tabular} & 0.2298 & 0.4330 (7.60) & \textbf{0.4970 (6.08)} \\
VGG16 (CIFAR100) & \begin{tabular}[c]{@{}l@{}}features{[}43{]} \\ (last CNN)\end{tabular} & 0.2616 & 0.5107 (7.58) & \textbf{0.5430 (6.07)} \\ \bottomrule
\end{tabular}
\caption{Average correlation score across neurons in different layers of VGG16(CIFAR100). We tested the second to last layer (features[4]) which is a fully connected layer, as well as the last convolutional layer features[43]. Number in brackets is average explanation length. Both layers reach quite high correlation scores, but require long explanations, indicating the neurons are polysemantic.}
\label{tab:vgg16_layers}
\end{table}

%% file: table/vit_layers.tex
\begin{table}[h!]
\centering
\begin{tabular}{@{}lllll@{}}
\toprule
Model & Target layer & CLIP-Dissect & LE(label) & LE(SigLIP) \\ \midrule
ViT-B/16 (ImageNet) & 11/11 - Residual Stream (end) & 0.0326 & 0.0813 & \textbf{0.1455} \\
ViT-B/16 (ImageNet) & 11/11 - MLP & 0.1722 & 0.3243 & \textbf{0.3489} \\
ViT-B/16 (ImageNet) & 10/11 - Residual Stream (end) & 0.0860 & 0.1293 & \textbf{0.2643} \\
ViT-B/16 (ImageNet) & 10/11 - MLP & 0.1067 & 0.1940 & \textbf{0.3088} \\ \bottomrule
\end{tabular}
\caption{Average simulation correlation scores of different methods for different layers of ViT-B-16. We can see MLP layers are more interpretable on average than the residual stream.}
\label{tab:vit_layers}
\end{table}

%% file: table/dead_neurons.tex
\begin{table}[h!]
\centering
\begin{tabular}{@{}lllllll@{}}
\toprule
Target model & \begin{tabular}[c]{@{}l@{}}ResNet-50\\ (ImageNet)\end{tabular} & \begin{tabular}[c]{@{}l@{}}ResNet-18\\ (Places-365)\end{tabular} & \begin{tabular}[c]{@{}l@{}}VGG-16\\ (CIFAR-100)\end{tabular} & \begin{tabular}[c]{@{}l@{}}ViT-B/16\\ (ImageNet)\end{tabular} & \begin{tabular}[c]{@{}l@{}}ViT-L/32\\ (ImageNet)\end{tabular} & \begin{tabular}[c]{@{}l@{}}ViT-L/16\\ (ImageNet)\end{tabular} \\ \midrule
Dead neurons: & 0\% & 0\% & 0\% & 5.27\% & 6.57\% & 100\% \\ \bottomrule
\end{tabular}
\caption{Fraction of dead neurons in different models.}
\label{tab:dead_neurons}
\end{table}

%% file: table/correlation_without_dead.tex
\begin{table}[h!]
\centering
\begin{tabular}{@{}llll@{}}
\toprule
Target model & CLIP-Dissect & LE (Label) & LE (SigLIP) \\ \midrule
ViT-B/16 (ImageNet) & 0.1807 +- 0.004 & 0.3399 +- 0.005 & \textbf{0.3629 +- 0.005} \\
ViT-L/32 (ImageNet) & 0.0590 +- 0.002 & 0.1984 +- 0.004 & \textbf{0.2254 +- 0.004} \\ \bottomrule
\end{tabular}
\caption{Average correlations scores of different explanation methods without dead neurons on ViT. We can see this improves average correlation scores by around 5\% over Table \ref{tab:correlation_score}}
\label{tab:correlation_no_dead}
\end{table}

%% file: example_paper.bbl
\begin{thebibliography}{35}
\providecommand{\natexlab}[1]{#1}
\providecommand{\url}[1]{\texttt{#1}}
\expandafter\ifx\csname urlstyle\endcsname\relax
  \providecommand{\doi}[1]{doi: #1}\else
  \providecommand{\doi}{doi: \begingroup \urlstyle{rm}\Url}\fi

\bibitem[Alain \& Bengio(2018)Alain and Bengio]{alain2018understanding}
Alain, G. and Bengio, Y.
\newblock Understanding intermediate layers using linear classifier probes, 2018.

\bibitem[Bai et~al.(2024)Bai, Iyer, Oikarinen, and Weng]{bai2024describeanddissect}
Bai, N., Iyer, R.~A., Oikarinen, T., and Weng, T.-W.
\newblock Describe-and-dissect: Interpreting neurons in vision networks with language models, 2024.

\bibitem[Bau et~al.(2017)Bau, Zhou, Khosla, Oliva, and Torralba]{netdissect2017}
Bau, D., Zhou, B., Khosla, A., Oliva, A., and Torralba, A.
\newblock Network dissection: Quantifying interpretability of deep visual representations.
\newblock In \emph{CVPR}, 2017.

\bibitem[Bau et~al.(2020)Bau, Zhu, Strobelt, Lapedriza, Zhou, and Torralba]{bau2020understanding}
Bau, D., Zhu, J.-Y., Strobelt, H., Lapedriza, A., Zhou, B., and Torralba, A.
\newblock Understanding the role of individual units in a deep neural network.
\newblock \emph{PNAS}, 2020.

\bibitem[Bills et~al.(2023)Bills, Cammarata, Mossing, Tillman, Gao, Goh, Sutskever, Leike, Wu, and Saunders]{bills2023language}
Bills, S., Cammarata, N., Mossing, D., Tillman, H., Gao, L., Goh, G., Sutskever, I., Leike, J., Wu, J., and Saunders, W.
\newblock Language models can explain neurons in language models.
\newblock \url{https://openaipublic.blob.core.windows.net/neuron-explainer/paper/index.html}, 2023.

\bibitem[Bricken et~al.(2023)Bricken, Templeton, Batson, Chen, Jermyn, Conerly, Turner, Anil, Denison, Askell, Lasenby, Wu, Kravec, Schiefer, Maxwell, Joseph, Hatfield-Dodds, Tamkin, Nguyen, McLean, Burke, Hume, Carter, Henighan, and Olah]{bricken2023monosemanticity}
Bricken, T., Templeton, A., Batson, J., Chen, B., Jermyn, A., Conerly, T., Turner, N., Anil, C., Denison, C., Askell, A., Lasenby, R., Wu, Y., Kravec, S., Schiefer, N., Maxwell, T., Joseph, N., Hatfield-Dodds, Z., Tamkin, A., Nguyen, K., McLean, B., Burke, J.~E., Hume, T., Carter, S., Henighan, T., and Olah, C.
\newblock Towards monosemanticity: Decomposing language models with dictionary learning.
\newblock \emph{Transformer Circuits Thread}, 2023.
\newblock https://transformer-circuits.pub/2023/monosemantic-features/index.html.

\bibitem[Bykov et~al.(2023)Bykov, Kopf, Nakajima, Kloft, and H{\"o}hne]{bykov2023labeling}
Bykov, K., Kopf, L., Nakajima, S., Kloft, M., and H{\"o}hne, M.~M.
\newblock Labeling neural representations with inverse recognition.
\newblock In \emph{NeurIPS}, 2023.

\bibitem[Cunningham et~al.(2023)Cunningham, Ewart, Riggs, Huben, and Sharkey]{cunningham2023sparse}
Cunningham, H., Ewart, A., Riggs, L., Huben, R., and Sharkey, L.
\newblock Sparse autoencoders find highly interpretable features in language models, 2023.

\bibitem[Elhage et~al.(2021)Elhage, Nanda, Olsson, Henighan, Joseph, Mann, Askell, Bai, Chen, Conerly, DasSarma, Drain, Ganguli, Hatfield-Dodds, Hernandez, Jones, Kernion, Lovitt, Ndousse, Amodei, Brown, Clark, Kaplan, McCandlish, and Olah]{elhage2021mathematical}
Elhage, N., Nanda, N., Olsson, C., Henighan, T., Joseph, N., Mann, B., Askell, A., Bai, Y., Chen, A., Conerly, T., DasSarma, N., Drain, D., Ganguli, D., Hatfield-Dodds, Z., Hernandez, D., Jones, A., Kernion, J., Lovitt, L., Ndousse, K., Amodei, D., Brown, T., Clark, J., Kaplan, J., McCandlish, S., and Olah, C.
\newblock A mathematical framework for transformer circuits.
\newblock \emph{Transformer Circuits Thread}, 2021.
\newblock https://transformer-circuits.pub/2021/framework/index.html.

\bibitem[Elhage et~al.(2023)Elhage, Lasenby, and Olah]{elhage2023priviliged}
Elhage, N., Lasenby, R., and Olah, C.
\newblock Privileged bases in the transformer residual stream, 2023.
\newblock URL \url{https://transformer-circuits.pub/2023/privileged-basis/}.

\bibitem[Erhan et~al.(2009)Erhan, Bengio, Courville, and Vincent]{erhan2009visualizing}
Erhan, D., Bengio, Y., Courville, A., and Vincent, P.
\newblock Visualizing higher-layer features of a deep network.
\newblock \emph{University of Montreal}, 1341\penalty0 (3):\penalty0 1, 2009.

\bibitem[Fong \& Vedaldi(2018)Fong and Vedaldi]{fong2018net2vec}
Fong, R. and Vedaldi, A.
\newblock Net2vec: Quantifying and explaining how concepts are encoded by filters in deep neural networks.
\newblock In \emph{ICCV}, 2018.

\bibitem[Geirhos et~al.(2023)Geirhos, Zimmermann, Bilodeau, Brendel, and Kim]{geirhos2023dont}
Geirhos, R., Zimmermann, R.~S., Bilodeau, B., Brendel, W., and Kim, B.
\newblock Don't trust your eyes: on the (un)reliability of feature visualizations, 2023.

\bibitem[Goh et~al.(2021)Goh, Cammarata, Voss, Carter, Petrov, Schubert, Radford, and Olah]{goh2021multimodal}
Goh, G., Cammarata, N., Voss, C., Carter, S., Petrov, M., Schubert, L., Radford, A., and Olah, C.
\newblock Multimodal neurons in artificial neural networks.
\newblock \emph{Distill}, 2021.
\newblock \doi{10.23915/distill.00030}.
\newblock https://distill.pub/2021/multimodal-neurons.

\bibitem[Guo et~al.(2017)Guo, Pleiss, Sun, and Weinberger]{guo2017calibration}
Guo, C., Pleiss, G., Sun, Y., and Weinberger, K.~Q.
\newblock On calibration of modern neural networks.
\newblock In \emph{ICML}, 2017.

\bibitem[Gurnee et~al.(2023)Gurnee, Nanda, Pauly, Harvey, Troitskii, and Bertsimas]{gurnee2023finding}
Gurnee, W., Nanda, N., Pauly, M., Harvey, K., Troitskii, D., and Bertsimas, D.
\newblock Finding neurons in a haystack: Case studies with sparse probing, 2023.

\bibitem[He et~al.(2016)He, Zhang, Ren, and Sun]{he2016deep}
He, K., Zhang, X., Ren, S., and Sun, J.
\newblock Deep residual learning for image recognition.
\newblock In \emph{CVPR}, 2016.

\bibitem[Hernandez et~al.(2022)Hernandez, Schwettmann, Bau, Bagashvili, Torralba, and Andreas]{hernandez2022natural}
Hernandez, E., Schwettmann, S., Bau, D., Bagashvili, T., Torralba, A., and Andreas, J.
\newblock Natural language descriptions of deep visual features.
\newblock In \emph{ICLR}, 2022.

\bibitem[Kalibhat et~al.(2023)Kalibhat, Bhardwaj, Bruss, Firooz, Sanjabi, and Feizi]{kalibhat2023identifying}
Kalibhat, N., Bhardwaj, S., Bruss, C.~B., Firooz, H., Sanjabi, M., and Feizi, S.
\newblock Identifying interpretable subspaces in image representations.
\newblock In \emph{ICML}, 2023.

\bibitem[Lee et~al.(2023)Lee, Oikarinen, Chatha, Chang, Chen, and Weng]{lee2023importance}
Lee, J., Oikarinen, T., Chatha, A., Chang, K.-C., Chen, Y., and Weng, T.-W.
\newblock The importance of prompt tuning for automated neuron explanations.
\newblock In \emph{NeurIPS ATTRIB Workshop}, 2023.

\bibitem[Miller(1995)]{miller1995wordnet}
Miller, G.~A.
\newblock Wordnet: a lexical database for english.
\newblock \emph{Communications of the ACM}, 38\penalty0 (11):\penalty0 39--41, 1995.

\bibitem[Mu \& Andreas(2020)Mu and Andreas]{mu2021compositional}
Mu, J. and Andreas, J.
\newblock Compositional explanations of neurons.
\newblock In \emph{NeurIPS}, 2020.

\bibitem[Nanfack et~al.(2023)Nanfack, Fulleringer, Marty, Eickenberg, and Belilovsky]{nanfack2023adversarial}
Nanfack, G., Fulleringer, A., Marty, J., Eickenberg, M., and Belilovsky, E.
\newblock Adversarial attacks on the interpretation of neuron activation maximization, 2023.

\bibitem[Oikarinen \& Weng(2023)Oikarinen and Weng]{oikarinen2023clip}
Oikarinen, T. and Weng, T.-W.
\newblock Clip-dissect: Automatic description of neuron representations in deep vision networks.
\newblock In \emph{ICLR}, 2023.

\bibitem[Olah et~al.(2020)Olah, Cammarata, Schubert, Goh, Petrov, and Carter]{olah2020zoom}
Olah, C., Cammarata, N., Schubert, L., Goh, G., Petrov, M., and Carter, S.
\newblock Zoom in: An introduction to circuits.
\newblock \emph{Distill}, 2020.
\newblock \doi{10.23915/distill.00024.001}.
\newblock https://distill.pub/2020/circuits/zoom-in.

\bibitem[Radford et~al.(2021)Radford, Kim, Hallacy, Ramesh, Goh, Agarwal, Sastry, Askell, Mishkin, Clark, Krueger, and Sutskever]{radford2021learning}
Radford, A., Kim, J.~W., Hallacy, C., Ramesh, A., Goh, G., Agarwal, S., Sastry, G., Askell, A., Mishkin, P., Clark, J., Krueger, G., and Sutskever, I.
\newblock Learning transferable visual models from natural language supervision, 2021.

\bibitem[Ribeiro et~al.(2016)Ribeiro, Singh, and Guestrin]{ribeiro2016should}
Ribeiro, M.~T., Singh, S., and Guestrin, C.
\newblock "why should i trust you?" explaining the predictions of any classifier.
\newblock In \emph{KDD}, 2016.

\bibitem[Rosa et~al.(2023)Rosa, Gilpin, and Capobianco]{larosa2023fuller}
Rosa, B.~L., Gilpin, L.~H., and Capobianco, R.
\newblock Towards a fuller understanding of neurons with clustered compositional explanations.
\newblock In \emph{NeurIPS}, 2023.

\bibitem[Sajjad et~al.(2022)Sajjad, Durrani, and Dalvi]{sajjad2022neuronlevel}
Sajjad, H., Durrani, N., and Dalvi, F.
\newblock Neuron-level interpretation of deep nlp models: A survey.
\newblock In \emph{TACL}, 2022.

\bibitem[Srivastava et~al.(2023)Srivastava, Oikarinen, and Weng]{srivastava2023corrupting}
Srivastava, D., Oikarinen, T., and Weng, T.-W.
\newblock Corrupting neuron explanations of deep visual features.
\newblock In \emph{ICCV}, 2023.

\bibitem[Wong et~al.(2021)Wong, Santurkar, and Madry]{wong2021leveraging}
Wong, E., Santurkar, S., and Madry, A.
\newblock Leveraging sparse linear layers for debuggable deep networks.
\newblock In \emph{ICML}, 2021.

\bibitem[Yu et~al.(2023)Yu, He, Deng, Shen, and Chen]{yu2023convolutions}
Yu, Q., He, J., Deng, X., Shen, X., and Chen, L.-C.
\newblock Convolutions die hard: Open-vocabulary segmentation with single frozen convolutional clip.
\newblock In \emph{NeurIPS}, 2023.

\bibitem[Zhai et~al.(2023)Zhai, Mustafa, Kolesnikov, and Beyer]{zhai2023sigmoid}
Zhai, X., Mustafa, B., Kolesnikov, A., and Beyer, L.
\newblock Sigmoid loss for language image pre-training.
\newblock In \emph{ICCV}, 2023.

\bibitem[Zhou et~al.(2015)Zhou, Khosla, Lapedriza, Oliva, and Torralba]{zhou2014object}
Zhou, B., Khosla, A., Lapedriza, A., Oliva, A., and Torralba, A.
\newblock Object detectors emerge in deep scene cnns.
\newblock In \emph{ICLR}, 2015.

\bibitem[Zimmermann et~al.(2023)Zimmermann, Klein, and Brendel]{zimmermann2023scale}
Zimmermann, R.~S., Klein, T., and Brendel, W.
\newblock Scale alone does not improve mechanistic interpretability in vision models.
\newblock In \emph{NeurIPS}, 2023.

\end{thebibliography}
